\documentclass{article}

\usepackage{microtype}
\usepackage{graphicx}
\usepackage{subcaption}
\usepackage{booktabs} 

\usepackage{hyperref}

\usepackage[accepted]{icml2026}
\makeatletter
\renewcommand{\Notice@String}{The $\mathit{3}^{rd}$ AI for Math Workshop at the $\mathit{43}^{rd}$ International Conference on Machine Learning (ICML), 2026. Copyright 2026 by the author(s).}
\renewcommand{\printAffiliationsAndNotice}[1]{\global\icml@noticeprintedtrue%
  {\let\thefootnote\relax\footnotetext{\hspace*{-\footnotesep}%
      \begin{minipage}[t]{0.94\columnwidth}
      \textsuperscript{1}Brown University\quad
      \textsuperscript{2}California Institute of Technology.\\
      Correspondence to: Jiayi Wu \texttt{<jiayi\_wu4@brown.edu>}.\\[0.35ex]
      \Notice@String
      \end{minipage}}}}
\makeatother

\usepackage{amsmath}
\usepackage{amssymb}
\usepackage{mathtools}
\usepackage{amsthm}

\usepackage[capitalize,noabbrev]{cleveref}

\usepackage[shortlabels]{enumitem}
\usepackage{placeins}
\usepackage{colortbl}          
\usepackage{fancyvrb}          
\usepackage[most,listings]{tcolorbox}   
\usepackage{listings}          

\usepackage{pifont}

\definecolor{itpblue}{RGB}{47,85,151}
\definecolor{itpborder}{RGB}{160,160,160}
\definecolor{itpbluebg}{RGB}{230,240,255}
\definecolor{itpgraybg}{RGB}{240,240,240}
\definecolor{itpbeige}{RGB}{255,248,230}

\newcommand{\cmark}{\textcolor{itpblue}{\ding{108}}}
\newcommand{\xmark}{\textcolor{gray}{\ding{109}}}
\newcommand{\pmark}{\textcolor{itpblue}{\ding{117}}}

\newtcolorbox{promptbox}[1][]{
  enhanced,
  colback=gray!5,
  colframe=itpborder,
  fonttitle=\small\bfseries,
  boxrule=0.4pt,
  arc=2pt,
  left=6pt, right=6pt, top=4pt, bottom=4pt,
  title={#1},
  breakable,
}
\newtcblisting{codebox}[1][]{
  enhanced,
  breakable,
  colback=white,
  colframe=itpblue,
  fonttitle=\small\bfseries,
  boxrule=0.6pt,
  arc=2pt,
  left=5pt, right=5pt, top=3pt, bottom=3pt,
  title={#1},
  listing only,
  listing options={
    basicstyle=\tiny\ttfamily,
    breaklines=true,
    breakatwhitespace=true,
    breakindent=0pt,
    postbreak=\mbox{\textcolor{gray}{$\hookrightarrow$}\space},
    columns=fullflexible,
    keepspaces=true,
    xleftmargin=0pt,
    xrightmargin=0pt,
  },
}

\theoremstyle{plain}

\theoremstyle{definition}

\theoremstyle{remark}

\usepackage[textsize=tiny]{todonotes}

\icmltitlerunning{\textsc{ITPEval}: Benchmarking Formal Translation Across Interactive Theorem Provers}

\begin{document}

\twocolumn[
  \icmltitle{\textsc{ITPEval}: Benchmarking Formal Translation Across \\ Interactive Theorem Provers}



  \icmlsetsymbol{equal}{*}

  \begin{icmlauthorlist}
    \icmlauthor{Jiayi Wu}{brown}
    \icmlauthor{Robert Joseph George}{cit}
    \icmlauthor{Anima Anandkumar}{cit}
  \end{icmlauthorlist}

  \icmlaffiliation{brown}{Brown University}
  \icmlaffiliation{cit}{California Institute of Technology}

  \icmlcorrespondingauthor{Jiayi Wu}{jiayi\_wu4@brown.edu}

  \icmlkeywords{Machine Learning, ICML}

  \vskip 0.3in
]



\printAffiliationsAndNotice{}  

\begin{abstract}
Formal theorem proving has emerged as a frontier challenge for machine learning, yet the ecosystem is fragmented: proofs remain siloed across incompatible systems, limiting both training data for learning-based provers and the portability of verified results. We present \textbf{\textsc{ITPEval}}, the first benchmark for evaluating automated formal proof translation across four major ITPs (Lean~4, Rocq, Isabelle, and HOL Light), spanning two distinct logical foundations. Our benchmark comprises 1{,}560 source files and 6{,}848 theorems organized into a \emph{controlled} tier of axiomatized files that isolates foundational translation difficulty, and an \emph{ecosystem} tier drawn from real libraries that exposes API and proof-style mismatches. We release \texttt{itpeval}, a unified multi-ITP verification infrastructure with state-isolated warm backends that preserve per-artifact native checking semantics. We evaluate both \emph{statement} and \emph{proof} translation across five frontier and open-weight LLMs on 12 directed translation pairs: statement translation peaks at 29.1\% pass@1 and proof translation at 10.5\%; controlled theorems reach 29.7\% proof pass@1 versus 5.2\% for ecosystem-level translations, indicating that library mismatch is the largest observed source of proof-translation failure. In addition to pass@$k$ evaluation, a deterministic Lean~4 BEq check establishes equivalence for 54.0\% of verified source $\mapsto$ Lean~4 miniF2F statement translations, showing that native type-checking alone can substantially overestimate semantic fidelity; an exploratory autoformalization/auto-informalization round-trip study suggests target-dependent verification patterns and possible benefits from multi-ITP context. Our benchmark\footnote{Data: \href{https://huggingface.co/datasets/jiayi005/ITPEval}{\texttt{jiayi005/ITPEval}}.}, verification infrastructure, and evaluation pipelines\footnote{Code: \href{https://github.com/lean-dojo/ITPEval}{\texttt{lean-dojo/ITPEval}}.} are publicly released.
\end{abstract} 

\section{Introduction}
\label{sec:intro}

\begin{figure*}[t]
    \centering
    \includegraphics[width=0.82\textwidth]{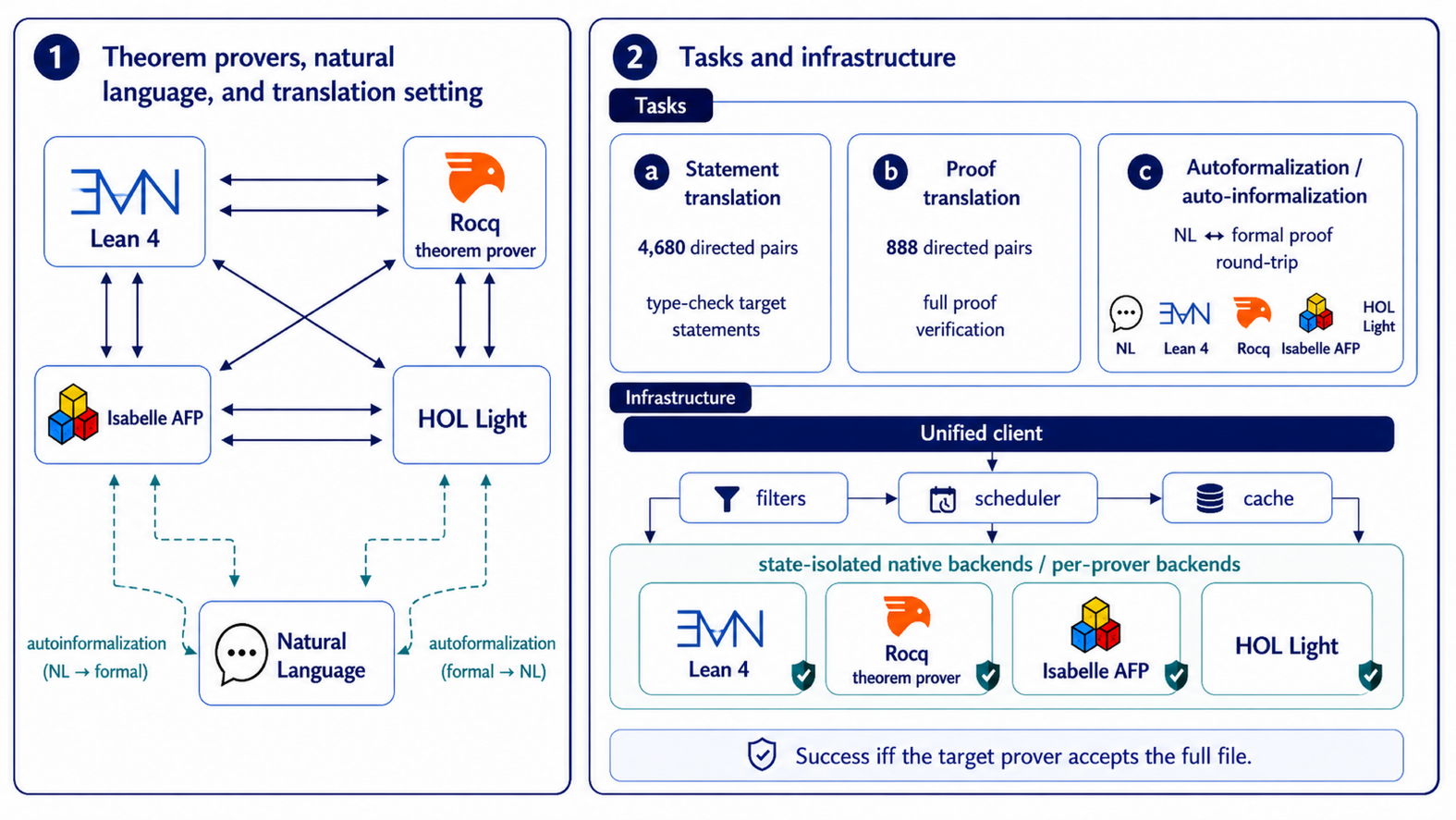}
    \caption{\textbf{Overview of ITPEval.} The figure summarizes the cross-ITP translation setting, natural-language round-trip tasks, and the unified state-isolated verification infrastructure.}
    \label{fig:placeholder}
\end{figure*}

Formal mathematics has become an increasingly important domain for AI research: large language models now achieve strong results on automated theorem proving~\citep{polu2022gptf,yang2024leandojo,first2023baldur,xin2024deepseekprover}, autoformalization~\citep{wu2022autoformalization,jiang2023dsp}, and mathematical reasoning more broadly. Yet while proving and autoformalization have been extensively researched and benchmarked, the formal ecosystem remains fragmented: each interactive theorem prover (ITP) implements its own logical foundation, tactic language, and mathematical library, and \emph{interoperability} between systems has been largely overlooked. 

This fragmentation brings various consequences: a theorem proved in Lean's Mathlib~\citep{mathlib2020} cannot be invoked in Isabelle's Archive of Formal Proofs~\citep{afp2004}, even when both encode the same mathematics. Cross-ITP translation, the task of converting formal proofs between systems while preserving correctness, would address this in two ways. First, it would let verified results transfer across ITP ecosystems, reducing duplicate formalization effort and expanding the training data available to learning-based provers. Second, it would inform autoformalization and auto-informalization, since both settings require preserving theorem meaning across different surface languages and libraries. We therefore also evaluate a round-trip autoformalization/auto-informalization setting in which multi-ITP context is used during generation.

Despite this importance, cross-ITP translation has received little systematic study. While formalization tracking efforts such as Formalizing 100 Theorems~\citep{wiedijk2008hundred} and the 1000+ Theorems project~\citep{thousand2024} catalog which classical results have been proved in which provers, and interoperability frameworks such as Dedukti~\citep{dowek2023dedukti,logipedia2020} enable proof exchange via shared intermediate representations, none of these provide evaluation benchmarks for measuring translation quality. Without standardized benchmarks, it is difficult to measure whether LLM-based or symbolic translation methods produce correct, idiomatic output, or to identify which system pairs and proof features remain out of reach.

Constructing such a benchmark is itself non-trivial: each
ITP community has developed its library largely independently, so no large aligned corpus of proofs exists across systems, and even theorems that have been formalized in multiple ITPs typically differ in proof structure, lemma dependencies, and library APIs. The only existing benchmark is Babel-formal~\citep{babel2025}, limited to 14 aligned files between Lean and Rocq. Evaluation methodology is itself an open problem: type-checking alone has a high false-positive rate for semantic correctness~\citep{poiroux2025reliable,liu2025gted}, and code translation benchmarks~\citep{yan2023codetransocean,zheng2023humaneval} usually apply only partially since ITP translation must navigate logical foundation differences, not just syntactic ones. These gaps motivate a benchmark that jointly addresses cross-system coverage and foundation-aware evaluation.

We present \textbf{\textsc{ITPEval}}, a benchmark and unified multi-ITP verification infrastructure for studying formal translation across major theorem-prover ecosystems. ITPEval spans Lean~4, Rocq, Isabelle, and HOL Light, covering both dependent type theory and higher-order logic. Its design separates two sources of difficulty: a \emph{controlled} tier of self-contained, axiomatized files that isolates foundational translation issues, and an \emph{ecosystem} tier of real library formalizations that exposes API mismatches, naming conventions, automation behavior, and proof-style differences. This tiered structure quantifies the additional cost of library and proof-engineering dependencies beyond foundation-level translation. Our contributions are:

\begin{itemize}[leftmargin=*]
    \item \textbf{A four-way aligned cross-ITP benchmark} with 1{,}560 source files and 6{,}848 theorems and lemmas across Lean~4, Rocq, Isabelle, and HOL Light. The benchmark yields 12 directed translation pairs and is organized into a controlled tier (64 axiomatized files, 660 lemmas) and an ecosystem tier (232 Formalizing 100 Theorems files and 1{,}264 miniF2F statements).

    \item \textbf{A unified multi-ITP client and verification infrastructure} that exposes heterogeneous theorem provers through a common evaluation interface with environment-keyed scheduling, state-isolated warm backends, and per-artifact diagnostics.

    \item \textbf{A systematic evaluation of closed-source and open-weight LLMs} on statement and proof translation. We evaluate GPT-5.5, Claude Sonnet~4.6, Gemini~3.1 Pro, DeepSeek-V4-Pro, and Qwen3-235B-A22B, with every output checked by the native target prover rather than by surface-level heuristics.

    \item \textbf{An exploratory autoformalization and auto-informalization round-trip study} showing target-prover effects and suggesting that multi-ITP context can affect downstream formalization success.

    \item \textbf{A Bidirectional Extended Definitional Equivalence check} (BEq) that tests whether verified translations preserve theorem meaning, implemented for Lean~4 targets. Applied to miniF2F statement translations, 174 of 322 verified source $\mapsto$ Lean~4 translations (54.0\%) pass BEq; the remaining 46.0\% expose either semantic drift or limitations of the restricted equivalence search.
\end{itemize}

\section{Related Work}
\label{sec:positioning}
\begin{table}[t]
\caption{\textbf{Positioning of \textsc{ITPEval}.}
A~= aligned data, B~= benchmark, N~= target-prover checking, U~= unified verifier.
\cmark~= central or supported, \pmark~= partial or limited, \xmark~= not a focus.}
\label{tab:positioning}
\centering
\scriptsize
\setlength{\tabcolsep}{3.4pt}
\renewcommand{\arraystretch}{1.06}
\resizebox{\linewidth}{!}{%
\begin{tabular}{>{\raggedright\arraybackslash}p{3.25cm}
                >{\raggedright\arraybackslash}p{2.25cm}
                >{\centering\arraybackslash}p{0.65cm}
                >{\centering\arraybackslash}p{0.65cm}
                >{\centering\arraybackslash}p{0.65cm}
                >{\centering\arraybackslash}p{0.65cm}
                >{\raggedright\arraybackslash}p{2.55cm}}
\toprule
\rowcolor{itpbluebg}
\textbf{Work} & \textbf{Systems} & \textbf{A} & \textbf{B} & \textbf{N} & \textbf{U} & \textbf{Role} \\
\midrule

\rowcolor{itpgraybg}
\multicolumn{7}{l}{\textbf{Interoperability and overlap}} \\
Formalizing 100 Theorems / 1000+ Theorems~\citep{wiedijk2008hundred,thousand2024}
& Many ITPs & \pmark & \xmark & \xmark & \xmark & Overlap catalogues \\

Dedukti / Logipedia / EuroProofNet~\citep{dowek2023dedukti,logipedia2020,thire2020interoperability}
& Many logics & \pmark & \xmark & \pmark & \pmark & Proof exchange \\

OpenTheory~\citep{opentheory2011}
& HOL family & \pmark & \xmark & \pmark & \pmark & HOL packages \\

\midrule
\rowcolor{itpgraybg}
\multicolumn{7}{l}{\textbf{Formal translation and autoformalization}} \\
Babel-formal~\citep{babel2025}
& Lean, Rocq & \cmark & \cmark & \cmark & \xmark & Two-prover benchmark \\

MiniF2F in Rocq~\citep{viennot2025minif2f}
& Lean, Isabelle, Rocq & \pmark & \pmark & \cmark & \xmark & Rocq statement port \\

MMA~\citep{jiang2023mma}
& Lean, Isabelle & \pmark & \xmark & \pmark & \xmark & NL--formal data \\

ProofWala~\citep{deshpande2025proofwala}
& Lean, Coq/Rocq & \pmark & \xmark & \cmark & \pmark & Proof search \\

KELPS~\citep{zhang2025kelps}
& Lean, Coq/Rocq, Isabelle & \pmark & \xmark & \pmark & \xmark & Autoformalization \\

\midrule
\rowcolor{itpgraybg}
\multicolumn{7}{l}{\textbf{Interfaces and analogues}} \\
LeanDojo / Pantograph / PISA / \texttt{itp-interface}~\citep{yang2024leandojo,anand2024pantograph,jiang2021pisa,itpinterface2024}
& One or two provers & \xmark & \xmark & \pmark & \pmark & Prover interfaces \\

CodeTransOcean / HumanEval-X / CRUXEval-X~\citep{yan2023codetransocean,zheng2023humaneval,xu2024cruxeval}
& Programming languages & \cmark & \cmark & \pmark & \xmark & Code translation \\

\midrule
\rowcolor{itpbeige}
\textbf{\textsc{ITPEval}}
& \textbf{Lean~4, Rocq, Isabelle, HOL Light}
& \textbf{\cmark} & \textbf{\cmark} & \textbf{\cmark} & \textbf{\cmark}
& \textbf{Four-way benchmark + verifier} \\

\bottomrule
\end{tabular}}
\end{table}

\paragraph{Formal systems.}
\textsc{ITPEval} spans two major proof-assistant traditions. Lean~4~\citep{demoura2021lean4} and Rocq~\citep{bertot2004coq} are based on the Calculus of Inductive Constructions (CIC), while Isabelle/HOL~\citep{nipkow2002isabelle} and HOL Light~\citep{harrison1996hol} are based on Higher-Order Logic (HOL). Translation difficulty is not determined by foundation alone: tactic languages, proof idioms, libraries, automation, and naming conventions differ even within the same logical family.

\paragraph{Prior work.}
Prior work has tracked theorem coverage across provers and built proof-exchange infrastructure, but has not provided a balanced four-way benchmark for cross-ITP translation. Formalization trackers record overlap across systems~\citep{wiedijk2008hundred,thousand2024}, while Dedukti, Logipedia, EuroProofNet, and OpenTheory provide proof exchange infrastructure~\citep{dowek2023dedukti,logipedia2020,thire2020interoperability,opentheory2011}. Babel-formal~\citep{babel2025} is the closest prior benchmark, but covers Lean and Rocq only. Recent multilingual autoformalization and proof-search systems study natural language to formal translation or two-prover proof search, rather than balanced formal-to-formal translation across four ITPs~\citep{jiang2023mma,deshpande2025proofwala,zhang2025kelps}.

\paragraph{Evaluation interfaces.}
Target-prover checking is necessary for formal translation, but type checking alone does not guarantee semantic fidelity~\citep{poiroux2025reliable,liu2025gted}. Existing interfaces such as LeanDojo, Pantograph, PISA, and \texttt{itp-interface} support interaction with individual provers or small prover sets~\citep{yang2024leandojo,anand2024pantograph,jiang2021pisa,itpinterface2024}. \textsc{ITPEval} instead contributes a unified, state-isolated multi-ITP verifier for benchmark-scale evaluation across Lean~4, Rocq, Isabelle, and HOL Light. Appendix~\ref{app:related-work} gives a more detailed comparison.


\section{Benchmark Design}
\label{sec:benchmark}

A benchmark for cross ITP translation must separate several sources of difficulty. Some failures come from differences in logical foundations, such as dependent type theory versus higher order logic. Others come from prover ecosystems: library APIs, naming conventions, automation, local helper lemmas, and proof idioms. \textsc{ITPEval} is designed to expose these effects separately, while evaluating both statement translation and full proof translation.

\paragraph{Tier A: controlled files.}
Tier~A contains self-contained, axiomatized files. These files include their own definitions and assumptions, and avoid dependence on prover-specific libraries. They are intended to isolate foundational and language-level translation issues, such as type theory, quantification, implicit arguments, universe conventions, and proof syntax.

\paragraph{Tier B: ecosystem files.}
Tier~B contains formalizations drawn from existing libraries and community developments. These files reflect real proof engineering practice: they use library lemmas, local helper definitions, tactic automation, naming conventions, and prover-specific mathematical structures. They therefore test the combined difficulty of translating both the mathematics and the surrounding ecosystem.

This two-tier design makes the main analysis interpretable. Performance on Tier~A measures how well models handle controlled cross-system translation, while the gap between Tier~A and Tier~B measures the additional cost of real prover ecosystems.

\subsection{Data Sources and Selection}
\label{sec:selection}

A central design constraint is the \textbf{four-way intersection requirement}: every file in the benchmark must be formalized in all four ITPs. This ensures that every source--target pair is evaluated on exactly the same theorems, enabling clean directional and foundation-pair comparisons without missing-data confounds. We exclude files where a formalization is unavailable in any of the four ITPs, where the source fails to compile in its native prover, or where theorem identity is ambiguous across systems.

\textbf{Babel-formal (Tier A, 16 aligned items; 64 files; 660 lemmas).} We adopt and extend the Babel-formal benchmark~\citep{babel2025} from its original 14-file Lean--Rocq scope to all four ITPs, re-formalizing each item in Isabelle and HOL Light and adding 2 new items formalized in all four systems from inception. Alignment is exact: we authored all four versions ourselves. Each item is a self-contained axiomatized theory (2--19 lemmas per prover-specific file) covering order theory, group theory, topology, and analysis. The axiomatized design isolates type-theoretic translation difficulty (dependent types, universe levels, implicit arguments) from library API mismatches.

\textbf{Formalizing 100 Theorems (Tier B, 58 aligned items; 232 files; 4{,}924 lemmas).} From Wiedijk's 100 classical theorems~\citep{wiedijk2008hundred}, we retain the 58 with formalizations publicly available in all four ITPs, aligned by canonical list number (e.g., Theorem~1 = irrationality of $\sqrt{2}$). Because each prover's formalization was developed independently by its community, stylistic differences are genuine. Formalizations are sourced from Mathlib~4 for Lean~4~\citep{mathlib2020}; the Rocq standard library, MathComp, and Rocq community libraries for Rocq~\citep{mathcomp2026,corn2026,coq100theorems2026,qarith2026,highschoolgeometry2026,coqtail2026,fsets2026,cauchyschwarz2017,coqfriendship2026,ekdohibscoqproofs2026}; Isabelle and the Archive of Formal Proofs for Isabelle/HOL~\citep{nipkow2002isabelle,afp2004}; and HOL Light's core theories for HOL Light~\citep{harrison1996hol}. Each translation unit is the \emph{entire source file}, including the headline theorem plus supporting lemmas and auxiliary definitions (mean 21 lemmas per file, range 1--95), reflecting realistic translation scenarios where proofs depend on locally-defined helpers.

\textbf{MiniF2F (Tier B, 1{,}264 single-theorem files, statements only).} The competition mathematics benchmark miniF2F~\citep{zheng2022minif2f} provides formalized problem statements across Lean, Isabelle, and HOL Light; we use the Rocq port of MiniF2F~\citep{viennot2025minif2f} and retain the 316 problems formalized in all four systems, aligned by standardized problem identifier (e.g., \texttt{amc12a\_2008\_p25}). Each file contains exactly one theorem statement. MiniF2F complements Formalizing 100 Theorems: its single-statement files test pure statement translation without multi-lemma structure, and its competition-math provenance provides diverse content (algebra, number theory, combinatorics) at controlled complexity. 

\subsection{Translation Tasks}

We evaluate two translation tasks, with \textbf{statement translation as the primary task}. Statement translation is both more tractable and directly diagnostic of whether models can encode the same theorem across type-theoretic and definitional boundaries, without confounding this with proof-search difficulty.

\textbf{Statement translation (\texttt{stmts}).} Given a source file with all proof bodies replaced by placeholders (e.g., \texttt{sorry}), produce the corresponding file in the target ITP, also with placeholder proofs. Verification checks that the generated file \emph{type-checks} in the target system; for multi-lemma files (Babel-formal, Formalizing 100 Theorems), all statements in the file must type-check together. 

\textbf{Proof translation (\texttt{proofs}).} Given a complete proof file in the source ITP, produce a complete, compilable proof file in the target ITP. Verification requires that the entire generated file compiles without errors and with no sorry placeholders; for multi-lemma modules, every lemma must be fully proved. 

\subsection{Infrastructure}
\label{sec:infra}

A substantial part of building \textsc{ITPEval} was designing a scalable verification infrastructure, not merely collecting aligned formalizations. Cross ITP evaluation is only meaningful if generated artifacts are checked by the target prover itself under conditions that faithfully reflect standalone verification. We refer to this as \emph{native target-prover checking}. For this reason, all labels in \textsc{ITPEval} are produced by the actual target ITP. A translation is counted as \emph{verified} only when the target prover accepts the entire generated file with exit code~0. Partial files, later declaration failures, and proof mode outputs containing placeholders are all counted as failures. For every artifact, the verifier records the final verdict, diagnostics, wall clock time, backend identity, and environment metadata.

The main systems challenge is that the target provers have very different execution models. Lean~4 and Rocq can be checked relatively directly, but Isabelle and HOL Light have much heavier startup and session costs. Isabelle verification is organized around sessions and theory files, while HOL Light runs through an OCaml-based environment whose initialization can dominate small checks. Treating each prover as a separate script would make evaluation fragile, difficult to restart, and hard to compare across targets. We therefore release \texttt{itpeval}, a unified multi-ITP client and toolchain manager. It installs and runs all four proof assistants behind a common \texttt{run\_itp()} entry point: each target prover is materialized in its own task directory, while prover-specific details remain localized to adapters.

\begin{figure*}[t]
  \centering
  \includegraphics[width=0.85\textwidth]{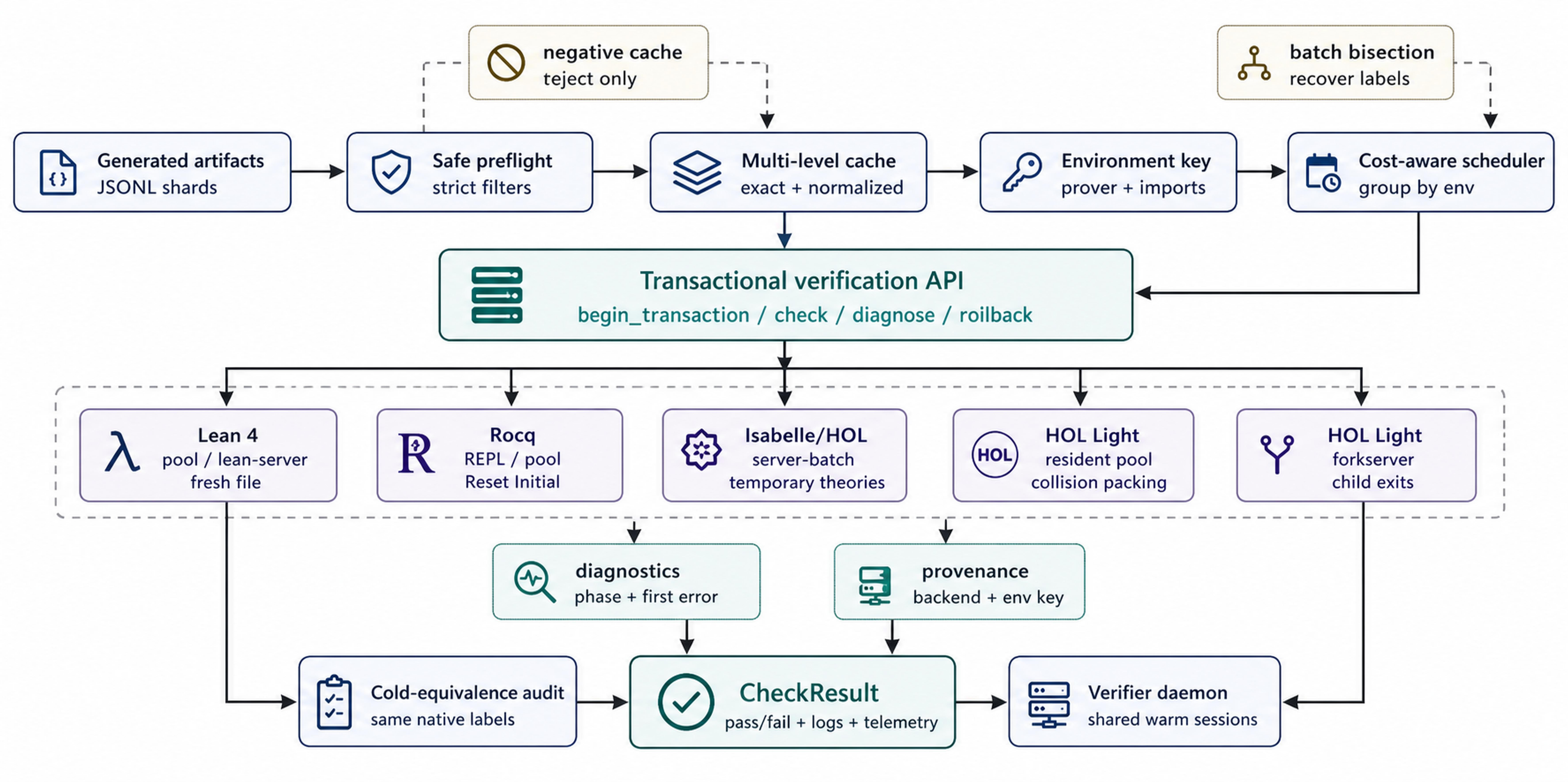}
  \caption{\textsc{ITPEval} verification infrastructure. Generated artifacts are filtered, cached, scheduled by environment, and checked by prover-specific warm backends while preserving native per-file checking semantics.}
  \label{fig:infra-overview-main}
\end{figure*}

The central invariant is \emph{state isolation}: every accelerated check must be observationally equivalent to verifying the generated artifact on its own. Figure~\ref{fig:infra-overview-main} summarizes the verifier. Generated JSONL records pass through strict one-sided filters, content-addressed caching, environment keys, and cost-aware scheduling before entering a transactional \texttt{CheckRequest}/\texttt{CheckResult} API. The API then dispatches to prover-specific backends: Lean~4 checks fresh synthetic files, Rocq resets persistent REPL workers with \texttt{Reset Initial.}, Isabelle checks temporary theories inside server sessions, and HOL Light uses collision-aware resident workers or a forkserver. Adaptive bisection recovers per-artifact labels after batch failures, append-only checkpoints make runs restartable, and telemetry records diagnostics, backend provenance, and timing. These mechanisms are part of the experimental protocol: without them, warm workers can leak declarations across files, batched checking can lose pass@1 labels, and target-prover effects can be confounded with verifier behavior.

\subsection{Semantic Equivalence Checking}
\label{sec:beq-design}

Target-prover type checking is necessary for correctness but not sufficient for semantic fidelity: a translated statement can type-check while expressing a weaker or shifted theorem~\citep{poiroux2025reliable,liu2025gted}. To quantify this gap, we adapt \emph{Bidirectional Extended Definitional Equivalence} (BEq), introduced by \citet{liu2025beq}, to verified Lean~4 target translations. Given a generated Lean statement $G$ and the reference Lean statement $R$, BEq checks both entailments $G \vdash R$ and $R \vdash G$; a translation passes BEq iff both directions succeed.

Each directional check constructs a fresh Lean file by renaming the assumed theorem to \texttt{stmt\_assumed}, renaming the goal theorem to \texttt{beq\_goal}, replacing the goal proof with restricted proof search, and running Lean. Liu et al.'s original implementation first tries \texttt{exact?} and then samples tactic sequences from an LLM under a restricted primitive set. We keep the same bidirectional criterion and restricted-primitive design, but replace LLM-sampled transformations with a deterministic Lean tactic cascade. This makes the metric reproducible and avoids adding a second model-dependent source of variance to the translation benchmark. Appendix~\ref{app:beq} gives the full protocol and failure analysis.


\section{Evaluation and Results}
\label{sec:eval}

\subsection{Evaluation Setup}
\label{sec:eval-setup}

We evaluate five zero-shot, greedy-decoded models ($T=0$): GPT-5.5, Claude Sonnet~4.6, Gemini~3.1 Pro Preview, DeepSeek-V4-Pro, and Qwen3-235B-A22B. This set covers closed-source frontier models and large open-weight models. Direct translation is the primary setting: given a source file and a target ITP, the model must emit only raw target ITP code, with no markdown fencing or explanation. We report pass@1 results: pass@1 is the fraction of translation instances for which the single generated candidate verifies under the target prover; more generally, pass@$k$ counts an instance as solved if any of $k$ generated candidates verifies.

\subsection{Verification Throughput}
\label{sec:infra-results}

The unified verifier is necessary in practice because target provers differ sharply in startup and checking cost. Table~\ref{tab:infra-backends-main} reports wall clock time on 16 Babel-formal statement files. Lean~4 and Rocq already have modest cold checking costs, but still benefit from pooled or persistent workers. The largest gains occur for Isabelle and HOL Light, where startup and environment initialization dominate naive checking. Isabelle decreases from 472.2\,s to 6.9\,s using server batch sessions, while HOL Light decreases from 527.4\,s to 149.0\,s using resident or forkserver backends. These measurements show that scalable cross ITP evaluation requires infrastructure that is both faster and preserves per-file checking semantics.

\begin{table}[t]
  \caption{State preserving verifier backends and observed throughput on 16 Babel-formal statement files. Times are wall clock seconds; lower is better.}
  \label{tab:infra-backends-main}
  \centering
  \scriptsize
  \setlength{\tabcolsep}{2pt}
  \begin{tabular}{llllrr}
    \toprule
    Prover & Backend & Reuse & Isolation & Cold & Warm \\
    \midrule
    Lean~4 & pool/server & compiler workers & fresh file & 15.1 & 3.7 \\
    Rocq & REPL pool & persistent proc. & \texttt{Reset} & 4.3 & 0.5 \\
    Isabelle & server batch & persistent session & temp.\ theories & 472.2 & 6.9 \\
    HOL Light & forkserver & preloaded OCaml & child exit & 527.4 & 149.0 \\
    \bottomrule
  \end{tabular}
\end{table}
\subsection{Statement Translation}
\label{sec:stmts-results}

Statement translation is the primary evaluation in \textsc{ITPEval}, assessed on all 4{,}680 directed pairs per model across the three benchmarks.

\begin{figure}[t]
  \centering
  \includegraphics[width=0.74\linewidth]{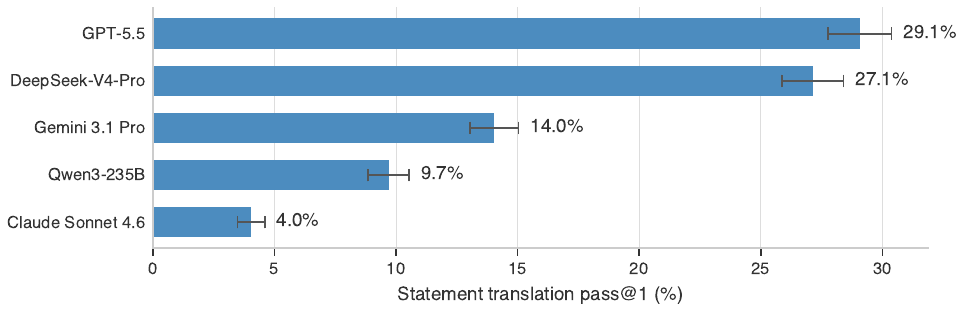}
  \caption{Statement translation pass@1 across all 4{,}680 directed file pairs. GPT-5.5 and DeepSeek-V4-Pro are the strongest overall, but even the best model succeeds on fewer than one third of pairs.}
  \label{fig:stmts-overall}
\end{figure}

Figure~\ref{fig:stmts-overall} shows that statement translation remains challenging even before proof search is required. GPT-5.5 achieves the highest overall statement pass rate (29.1\%), followed closely by DeepSeek-V4-Pro (27.1\%). Gemini is lower at 14.0\%, while Qwen3-235B-A22B and Claude Sonnet~4.6 remain below 10\%. The gap between the two leading models and the rest suggests that statement translation depends on both formal syntax control and target library familiarity.

\begin{table}[t]
  \caption{Statement translation pass@1 by benchmark. The Tier~A/B gap persists, but miniF2F outperforms Formalizing 100 Theorems despite containing competition mathematics.}
  \label{tab:stmts-bench}
  \centering
  \scriptsize
  \setlength{\tabcolsep}{2pt}
  \begin{tabular}{lrrr}
    \toprule
    Model & Babel (A, 192) & \shortstack{Formalizing 100\\Theorems (B, 696)} & miniF2F (B, 3792) \\
    \midrule
    GPT-5.5 & 91 (47.4\%) & 106 (15.2\%) & 1163 (30.7\%) \\
    Gemini 3.1 & 63 (32.8\%) & 97 (13.9\%) & 496 (13.1\%) \\
    Claude 4.6 & 21 (10.9\%) & 32 (4.6\%) & 134 (3.5\%) \\
    \midrule
    DeepSeek-V4 & 33 (17.2\%) & 46 (6.6\%) & 1190 (31.4\%) \\
    Qwen3-235B & 18 (9.4\%) & 10 (1.4\%) & 424 (11.2\%) \\
    \bottomrule
  \end{tabular}
\end{table}

Table~\ref{tab:stmts-bench} shows that the controlled tier is consistently easier than the ecosystem tier. Babel-formal removes most library dependence and isolates foundational translation, while Formalizing 100 Theorems requires real library APIs, local helper lemmas, naming conventions, and prover-specific mathematical structures. MiniF2F is often easier than Formalizing 100 Theorems because it contains single theorem statements with fewer local dependencies. This suggests that ecosystem complexity, not mathematical difficulty alone, is the dominant source of failure.

\subsubsection{Translation Direction Analysis}

Translation direction reveals strong asymmetry. In statement translation, Isabelle and HOL Light have the highest target-side verification rates, whereas Lean~4 and Rocq remain difficult. The gaps are large: Lean~4 $\mapsto$ Isabelle reaches 17.6\%, while Isabelle $\mapsto$ Lean~4 reaches only 4.3\%; HOL Light $\mapsto$ Isabelle reaches 25.8\%. These effects are not explained by logical foundation alone. CIC $\mapsto$ CIC directions are not consistently easier than cross-foundation directions, indicating that library conventions, elaboration behavior, and syntax matter more than the underlying foundation in many cases. Because every directed source-target pair is evaluated on the same mathematical content, these asymmetries reflect target-language generation difficulty rather than theorem selection. Full directional heatmaps are in Appendix~\ref{app:per-model}.

\subsection{Proof Translation}
\label{sec:proof-results}

We also evaluate \textbf{proof translation} on the 888 directed pairs from Babel-formal and Formalizing 100 Theorems.

\begin{table}[t]
  \caption{Proof translation pass@1 rates on 888 directed pairs. Tier~A is consistently much easier than Tier~B.}
  \label{tab:overall}
  \centering
  \scriptsize
  \setlength{\tabcolsep}{2.5pt}
  \begin{tabular}{lrrr}
    \toprule
    Model & Overall (888) & Babel (A, 192) & \shortstack{Formalizing 100\\Theorems (B, 696)} \\
    \midrule
    GPT-5.5 & \textbf{93 (10.5\%)} & 57 (29.7\%) & 36 (5.2\%) \\
    Gemini 3.1 & 45 (5.1\%) & 35 (18.2\%) & 10 (1.4\%) \\
    Claude 4.6 & 13 (1.5\%) & 11 (5.7\%) & 2 (0.3\%) \\
    \midrule
    DeepSeek-V4 & 13 (1.5\%) & 9 (4.7\%) & 4 (0.6\%) \\
    Qwen3-235B & 5 (0.6\%) & 3 (1.6\%) & 2 (0.3\%) \\
    \bottomrule
  \end{tabular}
\end{table}

Proof translation is much harder than statement translation. GPT-5.5 is the strongest model, but still solves only 10.5\% of proof pairs overall. The tier gap is also sharper: GPT-5.5 reaches 29.7\% on the controlled Babel-formal files but only 5.2\% on Formalizing 100 Theorems. This indicates that proof failure is not just a matter of producing syntactically valid target code. Complete proof translation often requires recovering target library dependencies, local helper lemmas, automation behavior, and prover-specific proof idioms.

\paragraph{Direction reversal.}
Proof translation has a different target profile from statement translation. Isabelle has one of the highest statement verification rates, but becomes the lowest proof target (1.9--3.2\%). HOL Light remains comparatively strong, with proof pass rates between 4.3\% and 5.7\%. This suggests that well-formed statement generation and complete proof construction reward different target-language properties. One interpretation is that Isabelle's structured Isar style helps models produce well-formed declarations, while complete proofs require more precise intermediate facts and automation control. Another possible interpretation is that HOL Light's compact OCaml tactic style appears more uniform for current models. Full per-model directional results are in Appendix~\ref{app:per-model}.

\subsection{Statements vs.\ Proofs}
\label{sec:stmts-vs-proofs}

On the 888 pairs where both modes are available, statement translation is clearly easier. A McNemar test on matched instances with the same theorem, direction, and model gives $\chi^2 = 249.19$ with $p < 0.0001$. The asymmetry is large: 417 instances pass only in statement mode, while 69 pass only in proof mode. The gap also varies by model. On the common benchmarks, GPT-5.5 drops from 22.2\% on statements to 10.5\% on proofs, Gemini from 18.0\% to 5.1\%, and Claude from 6.0\% to 1.5\%. At the theorem level, statement success is a near necessary condition for proof success: 32 theorems pass statements but never proofs, while no theorem passes proofs without any statement translation success. Thus, statement translation is a useful diagnostic for cross-system theorem encoding, but it is far from sufficient for full proof transfer.

\subsection{Aggregate Analysis}
\label{sec:aggregate}

We fit a logistic regression over all 27{,}840 evaluation records with binary verification outcome as the response ($\text{pseudo-}R^2 = 0.25$; full coefficients in Appendix~\ref{app:regression}). The strongest predictors are target prover and benchmark identity. HOL Light and Isabelle are strong statement targets, while Formalizing 100 Theorems and miniF2F are harder than Babel-formal. Translation task is also important: proof translation has a large negative effect ($-1.05$). By contrast, sharing a logical foundation has only a weak positive effect ($+0.19$, $p = 0.001$). In proof translation, all four foundation pairs cluster between 2.2\% and 4.5\%, consistent with ecosystem-level dependencies being the strongest measured source of difficulty once downstream prover libraries are involved.

Source complexity is another significant predictor. Longer source files are harder in proof mode ($-0.54$, $p < 0.001$), and source files associated with failed proof translations are longer than those associated with successful ones (median 3{,}992 vs.\ 1{,}978 characters). This is consistent with models struggling to translate long proofs into the target prover's local proof language, library names, and automation style.

\subsection{Autoformalization and Auto-informalization}
\label{sec:autoformalization}

Type-checking confirms syntactic well-formedness but not semantic fidelity: a generated statement can type-check while expressing a weaker or shifted theorem. To probe whether models preserve mathematical content through the natural-language interface, we evaluate round trips that alternate between autoformalization (NL $\mapsto$ formal) and auto-informalization (formal $\mapsto$ NL). Each round trip has three steps: formalize a natural-language theorem statement into ITP code, informalize the generated code back to natural language, and re-formalize the informalized description. Verification at the first and third steps is a coarse executable signal of whether the round trip preserves a formalizable theorem.

We evaluate two configurations on 100 miniF2F theorems across all five models:
\begin{itemize}[leftmargin=*]
    \item \textbf{Single-ITP round trip:} NL $\mapsto$ Lean~4 $\mapsto$ NL $\mapsto$ Lean~4.
    \item \textbf{Multi-ITP round trip:} NL $\mapsto$ \{Lean~4, Rocq, Isabelle, HOL Light\} $\mapsto$ NL $\mapsto$ \{Lean~4, Rocq, Isabelle, HOL Light\}.
\end{itemize}

\begin{table}[t]
  \caption{Multi-ITP autoformalization/auto-informalization round trips by target prover. Entries are pooled over five models; denominators exclude empty or skipped outputs.}
  \label{tab:autoformalization}
  \centering
  \footnotesize
  \setlength{\tabcolsep}{4pt}
  \begin{tabular}{lccc}
    \toprule
    Target & Step 1 pass & Step 3 pass & Both pass \\
    \midrule
    Lean~4    & 53/491 (10.8\%) & 56/493 (11.4\%) & 36 \\
    Isabelle  & 42/489 (8.6\%)  & 21/490 (4.3\%)  & 10 \\
    Rocq      & 152/465 (32.7\%) & 146/460 (31.7\%) & 101 \\
    HOL Light & 144/460 (31.3\%) & 147/454 (32.4\%) & 103 \\
    \bottomrule
  \end{tabular}
\end{table}

Table~\ref{tab:autoformalization} shows a descriptive target effect in the round-trip setting: Rocq and HOL Light verify about one third of pooled outputs at both formalization steps, while Lean~4 stays near 11\% and Isabelle drops to 4.3\% at step~3. The similar step-1 and step-3 rates for Rocq and HOL Light suggest that target-side differences can persist after the natural-language intermediate step. The benefit of multi-ITP context is model- and target-dependent: the strongest gains concentrate in a few model-target combinations, while other pairs show smaller or inconsistent changes. Thus, aligned multi-ITP context is promising evidence but not yet a uniform improvement across provers or models.

For Lean~4 specifically, multi-ITP context improves pooled verification relative to the single-ITP Lean cycle in this subset, with step~1 rising from 4.8\% to 10.6\%. We treat this as a hypothesis-generating result: larger samples and controlled ablations are needed before attributing the improvement to multi-ITP context rather than prompt, model, or theorem-selection effects.

\subsection{Semantic Equivalence Checking}
\label{sec:beq}

We apply the BEq check (Section~\ref{sec:beq-design}) to all 322 verified source $\mapsto$ Lean~4 statement translations on miniF2F.

\begin{table}[t]
  \caption{Semantic equivalence (BEq) of verified source $\mapsto$ Lean~4 statement translations on miniF2F. Pass\% is the fraction of verified translations whose equivalence to the reference is established by BEq.}
  \label{tab:beq}
  \centering
  \scriptsize
  \setlength{\tabcolsep}{3pt}
  \begin{tabular}{lrrrr}
    \toprule
    Model & Verified & BEq Pass & Pass\% & BEq Fail \\
    \midrule
    Claude Sonnet 4.6 &  74 &  62 &  83.8\% & 12 \\
    DeepSeek-V4-Pro  &  84 &  37 &  44.0\% & 47 \\
    Gemini 3.1 Pro   &   7 &   3 &  42.9\% & 4 \\
    GPT-5.5          &  29 &  10 &  34.5\% & 19 \\
    Qwen3-235B       & 128 &  62 &  48.4\% & 66 \\
    \midrule
    All models       & 322 & 174 &  54.0\% & 148 \\
    \bottomrule
  \end{tabular}
\end{table}

Table~\ref{tab:beq} shows that 174 of 322 verified translations pass BEq (54.0\%). This changes the interpretation of native verification: a Lean-accepted generated statement is often syntactically valid but not established equivalent to the reference by our restricted bidirectional check. Model differences are large. Claude Sonnet~4.6 has the highest BEq rate among verified translations (83.8\%), while GPT-5.5, Gemini 3.1 Pro, DeepSeek-V4-Pro, and Qwen3-235B range from 34.5\% to 48.4\%. BEq failures are mostly neither-direction failures (132 of 322 verified translations), while all asymmetric failures are backward-only (16 of 322): the reference entails the generated statement, but not conversely. This pattern is consistent with semantic weakening, though BEq failures should be read as \emph{not established equivalent} rather than definitive semantic inequivalence because the deterministic cascade is incomplete.

Source direction has a smaller effect than model choice, but the rates are not identical: Rocq $\mapsto$ Lean passes at 52.4\%, Isabelle $\mapsto$ Lean at 47.5\%, and HOL Light $\mapsto$ Lean at 61.5\%. HOL Light $\mapsto$ Lean has the highest BEq rate among verified translations, while Rocq $\mapsto$ Lean contributes the largest number of verified translations. These results show that target-prover type checking is necessary but weak as a semantic metric. It remains the executable acceptance signal for translation, but we need equivalence-aware evaluation to measure whether verified statements preserve theorem meaning.

\subsection{Discussion}
\label{sec:result-synthesis}
The largest performance gap is between controlled files and ecosystem files rather than between same-foundation and cross-foundation translation pairs. Models perform much better on the controlled tier, where library dependence is removed, and drop sharply on ecosystem files that require library APIs, naming conventions, automation, and local helper lemmas. Future translation systems should therefore prioritize retrieval, library mapping, API alignment, and proof-style adaptation. This parallels findings in programming language translation, where library-centric errors dominate over core language failures~\cite{xue2025translibeval}. Directional results reinforce this conclusion. If logical-foundation similarity were the main factor, Lean~4$\leftrightarrow$Rocq would have uniformly high verification rates; instead, verification rates vary substantially by target prover: Isabelle and HOL Light are strong statement targets, Isabelle becomes the lowest-verification proof target, and HOL Light remains relatively strong for proofs.

Because miniF2F, Formalizing 100 Theorems, and Babel-formal are public resources, absolute pass rates may be affected by training-data contamination; miniF2F also already has cross-ITP ports. We therefore emphasize controlled comparisons across targets, directions, tiers, and verification criteria under a fixed evaluation protocol; future private or newly authored splits would strengthen claims about out-of-distribution generalization.

Three evaluation limitations remain. BEq measures statement fidelity for Lean~4 targets, but semantic fidelity of proof strategies remains open: accepted proofs can establish equivalent statements while using different intermediate lemmas, automation, proof terms, or proof-engineering structure. The autoformalization/auto-informalization round-trip study is also exploratory and should be scaled up before drawing strong conclusions about multi-ITP context effects. Finally, our experiments evaluate zero-shot pass@1 for LLMs rather than pass@$k$ or agentic theorem-proving systems; agentic systems at the current frontier may achieve different performance levels and exhibit distinct failure patterns.

Finally, the infrastructure is part of the benchmark itself. Cross ITP results are only reliable when verifier state is isolated, labels are recovered per artifact, and backend provenance is recorded~\cite{xin2025apebench}. The unified state-isolated client therefore fixes the unit of measurement to one generated artifact while making large-scale target-prover checking reproducible and auditable.

\section{Conclusion}
\label{sec:conclusion}

We presented \textbf{\textsc{ITPEval}}, a benchmark and unified verification infrastructure for formal translation across Lean~4, Rocq, Isabelle, and HOL Light. Across 1{,}560 source files, 6{,}848 theorems, and five LLMs, statement translation peaks at 29.1\% pass@1 and proof translation at 10.5\%. The controlled-to-ecosystem drop identifies library dependencies as the strongest measured sources of translation difficulty. A BEq check establishes that type-checking alone can substantially overestimate semantic fidelity; the round-trip study suggests target-dependent autoformalization behavior and possible benefits from multi-ITP context. Limitations include zero-shot pass@1 evaluation and BEq coverage of Lean~4 targets only; future work should add retrieval and repair, deploy BEq beyond Lean~4, and scale aligned corpora.

\section*{Acknowledgments}
Jiayi Wu is supported by Caltech SURF fellowship. Robert Joseph George is supported by DARPA expMath and AA2.AIMING-1-NSF.2522494. Anima Anandkumar is supported in part by the Bren endowed chair, DARPA expMath, and the AI2050 Senior Fellowship program at Schmidt Sciences.

\bibliographystyle{icml2026}
\bibliography{references}

\clearpage
\appendix

\phantomsection
\section*{Appendix Overview}
\label{app:overview}

\begin{tcolorbox}[
  enhanced,
  colback=white,
  colframe=itpborder,
  colbacktitle=itpgraybg,
  coltitle=black,
  fonttitle=\bfseries,
  title={Appendix guide},
  boxrule=0.4pt,
  arc=2pt,
  left=8pt, right=8pt, top=6pt, bottom=6pt,
]
\begin{itemize}[leftmargin=*, itemsep=0.55em]
    \item \textbf{\hyperref[app:per-model]{Appendix~\ref*{app:per-model}: Full per-model results.}} Proof, statement, HOL Light target behavior, and tier-gap breakdowns by model.
    \item \textbf{\hyperref[app:statistics]{Appendix~\ref*{app:statistics}: Statistical analysis.}} Confidence intervals, model agreement, directional asymmetry, target rankings, foundation-pair effects, and variance decomposition.
    \item \textbf{\hyperref[app:data-complexity]{Appendix~\ref*{app:data-complexity}: Dataset and complexity characterization.}} Corpus sizes, source lengths, long-context proof translation, theorem difficulty, and miniF2F categories.
    \item \textbf{\hyperref[app:related-work]{Appendix~\ref*{app:related-work}: Extended related work.}} Formalization tracking, cross-ITP benchmarks, multilingual autoformalization, semantic fidelity, prover interfaces, and code-translation analogues.
    \item \textbf{\hyperref[app:experimental]{Appendix~\ref*{app:experimental}: Experimental details.}} Round-trip autoformalization results, prompt templates, and the Babel-formal formalization process.
    \item \textbf{\hyperref[app:beq]{Appendix~\ref*{app:beq}: Semantic equivalence checking details.}} BEq protocol, entailment asymmetries, and failure cases.
    \item \textbf{\hyperref[app:infra]{Appendix~\ref*{app:infra}: Infrastructure details.}} State-isolated verification architecture, telemetry, reproducibility, and released artifacts.
\end{itemize}
\end{tcolorbox}

\section{Full Per-Model Results}
\label{app:per-model}
\subsection{Proof Translation}

Figure~\ref{fig:direction-heatmaps-proof-permodel} shows per-model proof translation pass@1 by source--target direction.

\begin{center}
  \centering
  \includegraphics[width=\linewidth]{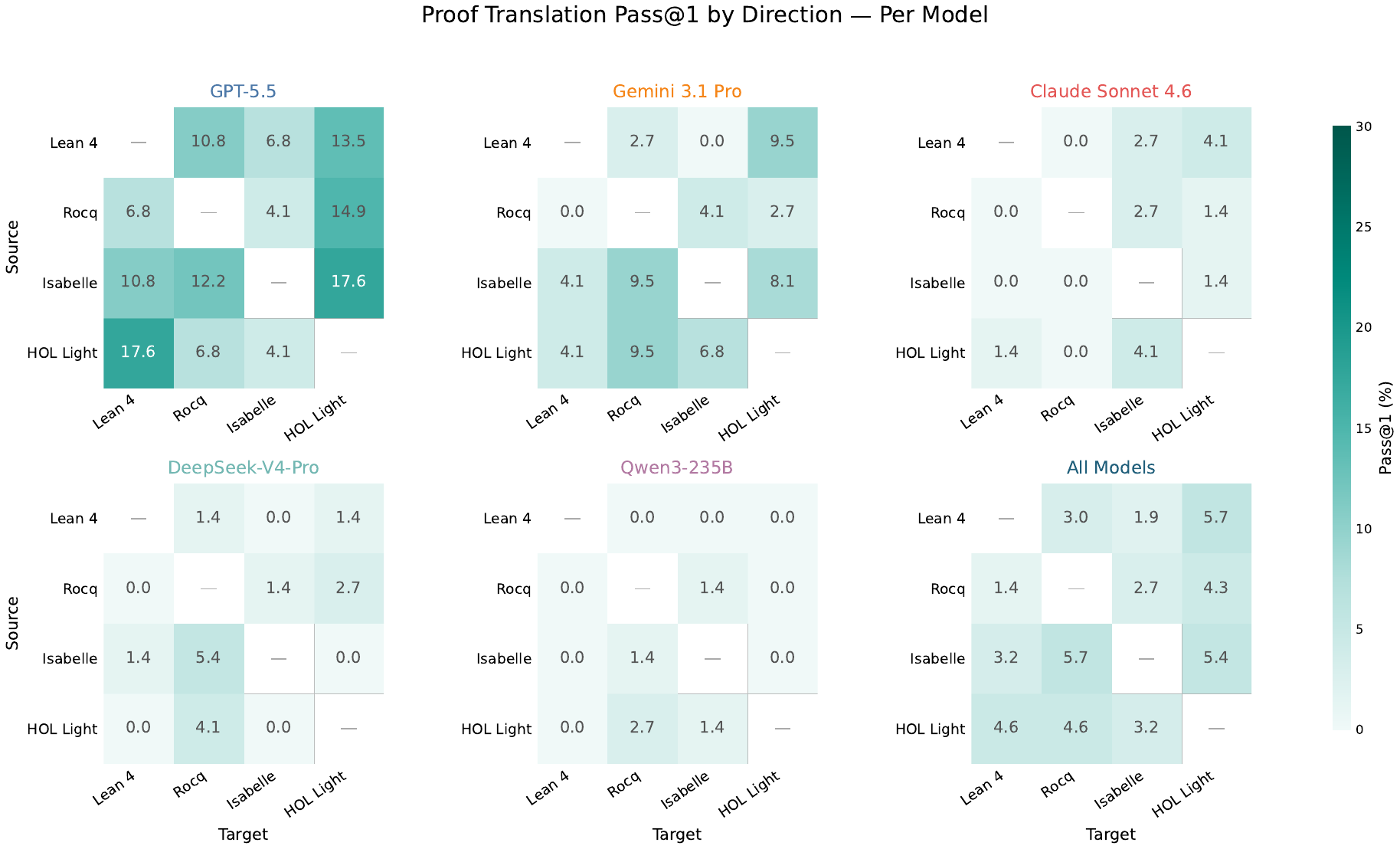}
  \captionof{figure}{Proof translation pass@1 (\%) by source--target direction, per model. GPT-5.5 dominates across most directions; DeepSeek-V4-Pro and Qwen3-235B show near-zero rates in many cells.}
  \label{fig:direction-heatmaps-proof-permodel}
\end{center}

\subsection{Statement Translation}

Figure~\ref{fig:direction-heatmaps-permodel} shows per-model statement translation pass@1 by source--target direction.

\begin{center}
  \centering
  \includegraphics[width=\linewidth]{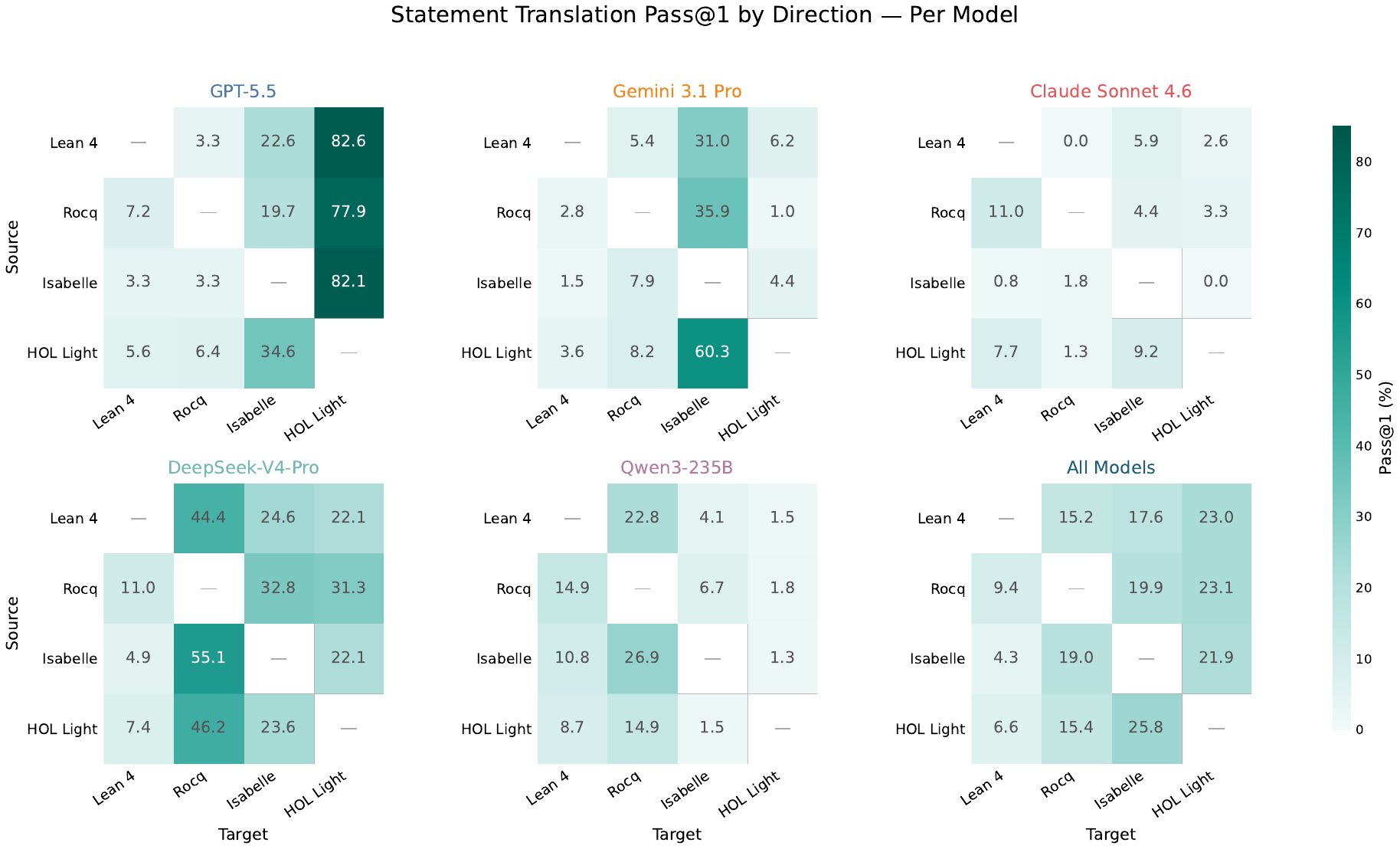}
  \captionof{figure}{Statement translation pass@1 (\%) by source--target direction, per model. Directional trends are broadly consistent but magnitudes vary: DeepSeek-V4-Pro achieves the highest per-cell rates while Claude Sonnet 4.6 is the weakest.}
  \label{fig:direction-heatmaps-permodel}
\end{center}

\FloatBarrier
\subsection{HOL Light as a Statement Target}
\label{app:hollight}

HOL Light-target statement verification is an order of magnitude more expensive than other targets, and the results are highly model-dependent. Table~\ref{tab:stmts-hollight} shows that GPT-5.5 achieves 80.9\% on HOL Light-target statements and DeepSeek-V4-Pro reaches 25.1\%, while the remaining three models stay below 4\%. This gap suggests that HOL Light statement generation is particularly sensitive to decoding stability and syntax discipline, rather than exclusively to mathematical content.

\begin{table}[!htbp]
  \caption{HOL Light-target statement translation by frontier model. Each direction has 390 pairs; Total has 1{,}170.}
  \label{tab:stmts-hollight}
  \centering
  \scriptsize
  \setlength{\tabcolsep}{2.5pt}
  \begin{tabular}{lcccc}
    \toprule
    Model & L4 $\mapsto$ HL & Rocq $\mapsto$ HL & Isa $\mapsto$ HL & Total \\
    \midrule
    GPT-5.5 & 322 (82.6\%) & 304 (77.9\%) & 320 (82.1\%) & 946 (80.9\%) \\
    Gemini 3.1 & 24 (6.2\%) & 4 (1.0\%) & 17 (4.4\%) & 45 (3.8\%) \\
    Claude 4.6 & 10 (2.6\%) & 13 (3.3\%) & 0 (0.0\%) & 23 (2.0\%) \\
    \bottomrule
  \end{tabular}
\end{table}

For Claude Sonnet~4.6, Isabelle $\mapsto$ HOL Light yields 0/390 passes, while Lean~4 $\mapsto$ HOL Light (2.6\%) and Rocq $\mapsto$ HOL Light (3.3\%) produce a handful. The dominant failure mode is syntax errors in HOL Light's OCaml-embedded tactic language: models frequently generate pseudo-HOL-Light that is syntactically close but fails the OCaml parser.

\FloatBarrier
\subsection{Tier Gap by Model}
\label{app:tier-gap}

Figure~\ref{fig:tier-gap} breaks down the Tier~A (controlled) vs.\ Tier~B (ecosystem) pass@1 gap for each model individually. The tier gap is consistent across all five models in both statement and proof translation, confirming that ecosystem complexity---library APIs, naming conventions, automation behavior, and local helper lemmas---is the dominant source of difficulty beyond foundational translation issues.

\begin{center}
  \centering
  \includegraphics[width=\linewidth]{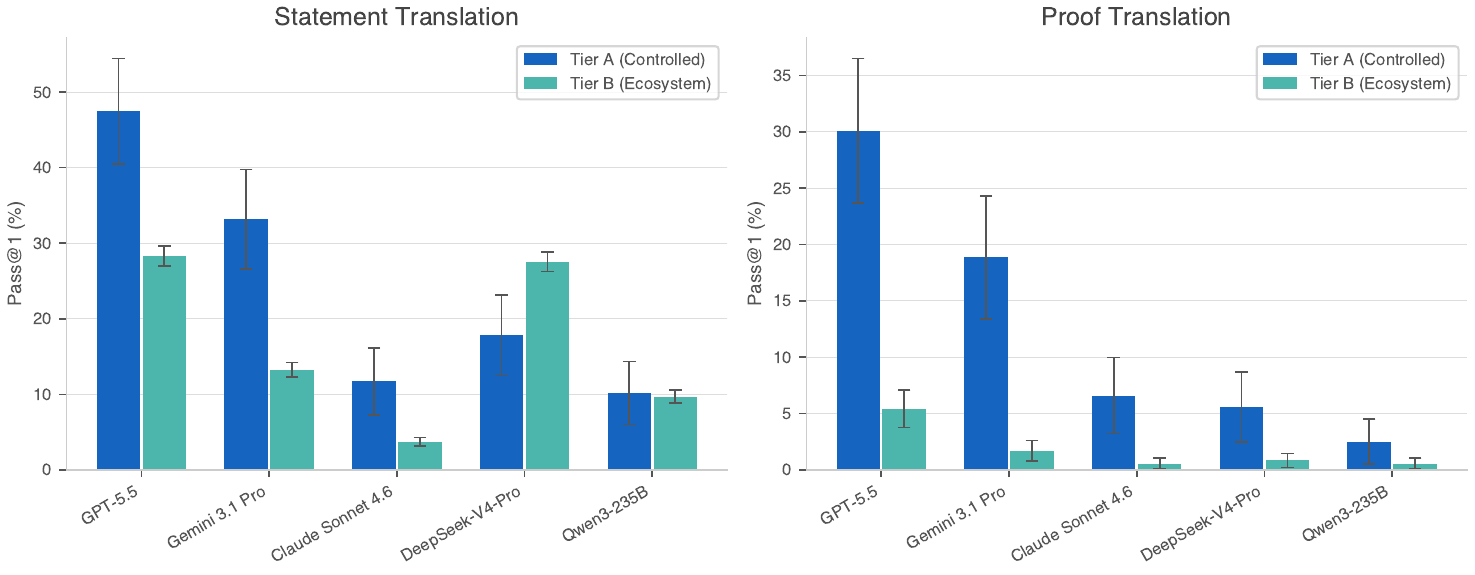}
  \captionof{figure}{Tier~A (controlled) vs.\ Tier~B (ecosystem) pass@1 rates by model. Left: statement translation; right: proof translation. The tier gap is consistent across all models: controlled, self-contained theorems are 3--10$\times$ more tractable than ecosystem-level translations with real library dependencies.}
  \label{fig:tier-gap}
\end{center}


\FloatBarrier
\section{Statistical Analysis}
\label{app:statistics}

\subsection{Confidence Intervals}
\label{app:ci}

Tables~\ref{tab:ci-stmts} and~\ref{tab:ci-proofs} augment the main-text pass@1 tables with 95\% Wilson confidence intervals. The narrow statement-mode intervals ($\pm$1--2\,pp) reflect the large sample size ($n = 4{,}680$); proof-mode intervals are wider ($\pm$2--4\,pp) due to the smaller evaluation set ($n = 888$).

\begin{table}[!htbp]
  \caption{Statement translation pass@1 with 95\% Wilson confidence intervals ($n = 4{,}680$ per model).}
  \label{tab:ci-stmts}
  \centering
  \small
  \begin{tabular}{lccl}
    \toprule
    Model & Passed & Pass@1 & 95\% CI \\
    \midrule
    GPT-5.5              & 1{,}360 & 29.1\% & [27.8, 30.4] \\
    DeepSeek-V4-Pro      & 1{,}269 & 27.1\% & [25.9, 28.4] \\
    Gemini 3.1 Pro       &    656  & 14.0\% & [13.1, 15.0] \\
    Qwen3-235B-A22B      &    452  &  9.7\% & [\phantom{0}8.8, 10.5] \\
    Claude Sonnet 4.6    &    187  &  4.0\% & [\phantom{0}3.5, \phantom{0}4.6] \\
    \bottomrule
  \end{tabular}
\end{table}

\begin{table}[!htbp]
  \caption{Proof translation pass@1 with 95\% Wilson confidence intervals ($n = 888$ per model).}
  \label{tab:ci-proofs}
  \centering
  \small
  \begin{tabular}{lccl}
    \toprule
    Model & Passed & Pass@1 & 95\% CI \\
    \midrule
    GPT-5.5              & 93 & 10.5\% & [\phantom{0}8.6, 12.7] \\
    Gemini 3.1 Pro       & 45 &  5.1\% & [\phantom{0}3.8, \phantom{0}6.7] \\
    DeepSeek-V4-Pro      & 13 &  1.5\% & [\phantom{0}0.9, \phantom{0}2.5] \\
    Claude Sonnet 4.6    & 13 &  1.5\% & [\phantom{0}0.9, \phantom{0}2.5] \\
    Qwen3-235B-A22B      &  5 &  0.6\% & [\phantom{0}0.2, \phantom{0}1.3] \\
    \bottomrule
  \end{tabular}
\end{table}

\subsection{Inter-Model Agreement}
\label{app:agreement}

To assess whether models fail on the same theorems or on complementary subsets, we compute Fleiss' $\kappa$ across all five models for each (theorem, direction) pair. The kappa values are very low: $\kappa = 0.032$ for statements ($n = 4{,}680$) and $\kappa = 0.108$ for proofs ($n = 888$). Both indicate only slight agreement beyond chance. Models largely fail on different theorems, suggesting that ensembling or best-of-$k$ strategies across models could yield higher pass rates than any single model. The higher proof-mode agreement reflects floor effects: with most models below 2\% pass rate, almost all entries are failures.

\subsection{Directional Asymmetry}
\label{app:asymmetry}

Table~\ref{tab:asymmetry} reports the full set of directional asymmetries for all six undirected ITP pairs. For each pair (A, B), we report the pooled pass@1 in both directions and the absolute gap. Statement-mode asymmetries are large and systematic: the strongest is Lean~4~$\leftrightarrow$~HOL Light ($|{\Delta}| = 16.4\,\text{pp}$), driven by higher verification rates for HOL Light targets. Proof-mode asymmetries are smaller, consistent with floor effects compressing differences.

\begin{table}[!htbp]
  \caption{Directional asymmetry: A $\mapsto$ B vs.\ B $\mapsto$ A pass@1 (\%), pooled across all models. Pairs are sorted by statement-mode gap.}
  \label{tab:asymmetry}
  \centering
  \footnotesize
  \setlength{\tabcolsep}{2.5pt}
  \begin{tabular}{lccclccc}
    \toprule
    & \multicolumn{3}{c}{Statements} & & \multicolumn{3}{c}{Proofs} \\
    \cmidrule(lr){2-4} \cmidrule(lr){6-8}
    Pair & A $\mapsto$ B & B $\mapsto$ A & $|\Delta|$ & & A $\mapsto$ B & B $\mapsto$ A & $|\Delta|$ \\
    \midrule
    L4$\leftrightarrow$HL   & 23.0 &  6.6 & 16.4 & & 5.7 & 4.6 & 1.1 \\
    L4$\leftrightarrow$Isa  & 17.6 &  4.3 & 13.4 & & 1.9 & 3.2 & 1.4 \\
    Rocq$\leftrightarrow$HL & 23.1 & 15.4 &  7.7 & & 4.3 & 4.6 & 0.3 \\
    L4$\leftrightarrow$Rocq & 15.2 &  9.4 &  5.8 & & 3.0 & 1.4 & 1.6 \\
    Isa$\leftrightarrow$HL  & 21.9 & 25.8 &  3.9 & & 5.4 & 3.2 & 2.2 \\
    Rocq$\leftrightarrow$Isa & 19.9 & 19.0 &  0.9 & & 2.7 & 5.7 & 3.0 \\
    \bottomrule
  \end{tabular}
\end{table}

The most symmetric statement pair is Rocq~$\leftrightarrow$~Isabelle ($|{\Delta}| = 0.9\,\text{pp}$), despite spanning different logical foundations. This suggests that the translation difficulty between these two systems is roughly balanced: neither is much easier to read from or generate into than the other. In proof mode, the largest asymmetry is Rocq~$\leftrightarrow$~Isabelle ($|{\Delta}| = 3.0\,\text{pp}$), where Isabelle $\mapsto$ Rocq (5.7\%) outperforms the reverse. Isabelle's verbose Isar proofs may provide richer structural cues for generating Rocq tactics.

\subsection{Bradley-Terry Target Ranking}
\label{app:bt}

To obtain a principled ranking of ITPs as generation targets, we fit a Bradley-Terry model to pairwise ``wins'': for each (theorem, model, source) triple, the target that the model successfully translates to ``beats'' targets that fail. Table~\ref{tab:bt} reports the log-strength parameters (mean-centered).

\begin{table}[!htbp]
  \caption{Bradley-Terry log-strength scores for ITPs as generation targets. Higher scores indicate higher target-side verification rates. The ranking reverses between modes: Isabelle drops from the second-highest statement target score to the lowest proof target score.}
  \label{tab:bt}
  \centering
  \small
  \begin{tabular}{lrr}
    \toprule
    Target ITP & Stmts score & Proofs score \\
    \midrule
    HOL Light  & $+0.51$ & $+0.55$ \\
    Isabelle   & $+0.35$ & $-0.35$ \\
    Rocq       & $+0.15$ & $+0.16$ \\
    Lean~4     & $-1.01$ & $-0.35$ \\
    \bottomrule
  \end{tabular}
\end{table}

HOL Light is the strongest target in both modes. The most striking shift is Isabelle: second-highest for statements ($+0.35$) but tied for lowest in proofs ($-0.35$). One possible explanation is that Isabelle's Isar language helps models generate well-formed declarations but is harder for constructing complete proofs. The gap between Lean~4 and the field narrows in proof mode, suggesting that Lean~4's difficulty as a statement target is partly driven by syntactic strictness rather than proof-search hardness.

\subsection{Logical Foundation Pairs}
\label{app:foundation}

Table~\ref{tab:foundation-pairs} breaks down pass rates by the logical foundation of the source and target provers (CIC-based: Lean~4, Rocq; HOL-based: Isabelle, HOL Light).

\begin{table}[!htbp]
  \caption{Pass@1 by logical foundation pair. In statement mode, the high CIC $\mapsto$ HOL and HOL $\mapsto$ HOL rates are driven by target-prover effects (Isabelle and HOL Light). In proof mode, all four pairs cluster within 2 percentage points.}
  \label{tab:foundation-pairs}
  \centering
  \footnotesize
  \setlength{\tabcolsep}{3pt}
  \begin{tabular}{llrrrr}
    \toprule
    & & \multicolumn{2}{c}{Statements} & \multicolumn{2}{c}{Proofs} \\
    \cmidrule(lr){3-4} \cmidrule(lr){5-6}
    Pair & Type & Pass/Total & \% & Pass/Total & \% \\
    \midrule
    CIC $\mapsto$ CIC & within & 479/3900 & 12.3 & 16/740 & 2.2 \\
    CIC $\mapsto$ HOL & cross  & 1630/7800 & 20.9 & 54/1480 & 3.6 \\
    HOL $\mapsto$ CIC & cross  & 883/7800 & 11.3 & 67/1480 & 4.5 \\
    HOL $\mapsto$ HOL & within & 932/3900 & 23.9 & 32/740 & 4.3 \\
    \midrule
    Within & & 1411/7800 & 18.1 & 48/1480 & 3.2 \\
    Cross  & & 2513/15600 & 16.1 & 121/2960 & 4.1 \\
    \bottomrule
  \end{tabular}
\end{table}

The statement-mode asymmetry between CIC $\mapsto$ HOL (20.9\%) and HOL $\mapsto$ CIC (11.3\%) is not a foundation effect but a target-prover effect: the HOL targets are Isabelle and HOL Light, while the CIC targets are Lean~4 and Rocq (both hard). In proof mode, the four foundation pairs cluster tightly (2.2--4.5\%), confirming that logical foundation is not a meaningful predictor of translation difficulty once other factors are controlled.

\subsection{Variance Decomposition}
\label{app:regression}

Table~\ref{tab:regression} reports the full logistic regression results from the main-effects model described in Section~\ref{sec:aggregate}. Reference categories are GPT-5.5 (model), statements (mode), Babel-formal (benchmark), and Lean~4 (source and target prover).

\begin{table}[!htbp]
  \caption{Logistic regression coefficients for verification outcome ($n = 27{,}840$, pseudo-$R^2 = 0.25$). Reference categories are GPT-5.5, statements, Babel-formal, Lean~4 source, and Lean~4 target.}
  \label{tab:regression}
  \centering
  \scriptsize
  \setlength{\tabcolsep}{2pt}
  \begin{tabular}{lrrcl}
    \toprule
    Predictor & Coef. & 95\% CI & $p$ & Note \\
    \midrule
    Intercept & $+2.25$ & $[1.6, 2.9]$ & $<.001$ & \\
    Tgt: HL (vs.\ L4) & $+2.34$ & $[2.2, 2.5]$ & $<.001$ & Strong tgt \\
    Tgt: Isa (vs.\ L4) & $+1.96$ & $[1.8, 2.1]$ & $<.001$ & Strong tgt \\
    Mod: DeepSeek (vs.\ GPT) & --- & --- & --- & \\
    Mod: Qwen (vs.\ GPT) & --- & --- & --- & \\
    Src: HL (vs.\ L4) & $+0.24$ & $[.09, .4]$ & .002 & Good src \\
    Same foundation & $+0.19$ & $[.07, .3]$ & .001 & Weak \\
    Src complexity (log len) & $-0.42$ & $[-.5, -.3]$ & $<.001$ & Harder \\
    Mod: Gemini (vs.\ GPT) & $-0.89$ & $[-1.0, -.8]$ & $<.001$ & \\
    Mode: proofs (vs.\ stmts) & $-1.05$ & $[-1.3, -.8]$ & $<.001$ & Harder \\
    Bench: F100 (vs.\ Babel) & $-1.50$ & $[-1.7, -1.3]$ & $<.001$ & Tier B \\
    Bench: miniF2F (vs.\ Babel) & $-2.33$ & $[-2.6, -2.1]$ & $<.001$ & Harder \\
    Mod: Claude (vs.\ GPT) & $-2.39$ & $[-2.5, -2.2]$ & $<.001$ & \\
    \midrule
    Src: Rocq (vs.\ L4) & $-0.06$ & $[-.2, .06]$ & .30 & n.s. \\
    Src: Isa (vs.\ L4) & $-0.13$ & $[-.3, .03]$ & .11 & n.s. \\
    Tgt: Rocq (vs.\ L4) & $-0.15$ & $[-.4, .05]$ & .13 & n.s. \\
    \bottomrule
  \end{tabular}
\end{table}

An interaction model adding model$\times$mode and model$\times$target interactions improves fit ($\text{pseudo-}R^2 = 0.26$), confirming that models have different relative strengths across modes and target provers.

\subsection{Statements vs.\ Proofs Correlation}
\label{app:stmts-proofs-corr}

At the theorem level, statement and proof translation difficulty are strongly correlated. Across the 74 theorems where both modes are evaluated (Babel-formal + Formalizing 100 Theorems), the Pearson correlation between per-theorem statement pass rate and proof pass rate is $r = 0.74$ ($p < 0.001$). Crucially, the relationship is asymmetric: all 41 theorems with at least one proof pass also have at least one statement pass, while 32 theorems pass statements but never proofs. No theorem passes proofs but fails all statement translations. This confirms that statement translation success is a near-necessary condition for proof translation success.

\FloatBarrier
\section{Dataset and Complexity Characterization}
\label{app:data-complexity}

\subsection{Dataset Statistics}
\label{app:dataset-stats}

Table~\ref{tab:dataset-stats} reports corpus-level statistics for the benchmark data. File lengths vary by orders of magnitude across benchmarks and ITPs: miniF2F files are short single-theorem statements (median 202 chars), while Formalizing 100 Theorems proof files range from $\sim$1\,KB to $\sim$750\,KB, with HOL Light files being the longest due to its verbose OCaml-embedded tactic style.

These proof-file lengths make Formalizing 100 Theorems an end-to-end long-context translation setting: we pass the full source file to the model without chunking or retrieval. Thus, proof translation on this tier reflects both formal translation ability and each model's effective long-context retention; we do not normalize results by context-window size or per-session memory behavior.

\begin{table}[!htbp]
  \caption{Source file length statistics (characters) by benchmark and mode. Per-ITP lengths are averaged across all four provers.}
  \label{tab:dataset-stats}
  \centering
  \footnotesize
  \setlength{\tabcolsep}{3pt}
  \begin{tabular}{llrrrr}
    \toprule
    Benchmark & Mode & Files & Mean & Median & Max \\
    \midrule
    Babel-formal     & stmts  &  16 &  3674 &  2744 &  11662 \\
    Babel-formal     & proofs &  16 &  6240 &  5118 &  21262 \\
    \shortstack[l]{Formalizing 100\\Theorems} & stmts  &  58 & 16711 &  6016 & 289797 \\
    \shortstack[l]{Formalizing 100\\Theorems} & proofs &  58 & 47543 & 16663 & 753999 \\
    miniF2F          & stmts  & 316 &   213 &   202 &   1499 \\
    \bottomrule
  \end{tabular}
\end{table}

Table~\ref{tab:dataset-itp-lens} breaks down statement file lengths by ITP. Isabelle and HOL Light files are consistently longer than their Lean~4 and Rocq counterparts, reflecting more verbose surface syntax. This length asymmetry is a surface-syntax confound when comparing target difficulty across ITPs.

\begin{table}[htbp]
  \caption{Median statement file length (characters) by ITP and benchmark.}
  \label{tab:dataset-itp-lens}
  \centering
  \footnotesize
  \setlength{\tabcolsep}{3pt}
  \begin{tabular}{lrrrr}
    \toprule
    Benchmark & Lean~4 & Rocq & Isabelle & HOL Light \\
    \midrule
    Babel-formal & 2536 & 2593 &  3254 &  3434 \\
    \shortstack[l]{Formalizing 100\\Theorems} & 6467 & 4733 &  8525 &  4566 \\
    miniF2F      &  207 &  192 &   270 &   129 \\
    \bottomrule
  \end{tabular}
\end{table}

\subsection{Source Complexity}
\label{app:complexity}

Source content length serves as a proxy for proof complexity. In proof mode, longer source proofs are significantly harder to translate (logistic regression coefficient $= -0.54$, $p < 0.001$). In statement mode, the effect is weaker but still negative (coefficient $= -0.10$, $p < 0.001$). Additionally, \emph{failed} proof translations are longer than successful ones (median 3{,}992 vs.\ 1{,}978 characters), suggesting that models over-generate when unable to find a correct proof. This is consistent with the general finding that LLMs struggle with long-context generation tasks and tend to produce verbose, incorrect outputs when the search space is large.

\begin{center}
  \centering
  \includegraphics[width=\linewidth]{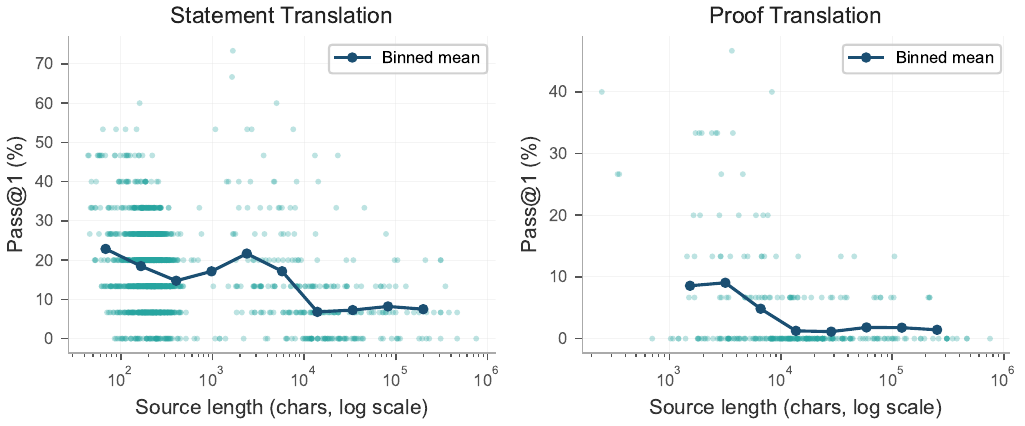}
  \captionof{figure}{Per-file pass rate vs.\ source content length (log scale). Left: statement translation; right: proof translation. Dark markers show binned means. The negative trend is stronger for proofs, where each doubling of source length roughly halves the success probability.}
  \label{fig:source-complexity}
\end{center}

\subsection{Theorem-Level Difficulty Distributions}
\label{app:theorem-difficulty}

Table~\ref{tab:theorem-dist} reports the distribution of per-file pass rates (aggregated across all models, directions, and modes) for each benchmark.

\begin{table}[!htbp]
  \caption{Distribution of file-level pass rates by benchmark. Pass rate is computed per file across all models and translation directions. No file exceeds 50\% in any benchmark.}
  \label{tab:theorem-dist}
  \centering
  \footnotesize
  \setlength{\tabcolsep}{3pt}
  \begin{tabular}{lrrccc}
    \toprule
    Benchmark & Files & Mean & 0\% & 0--10\% & 10--50\% \\
    \midrule
    Babel-formal & 16 & 17.8\% & 0 (0\%) & 2 (12\%) & 14 (88\%) \\
    \shortstack[l]{Formalizing 100\\Theorems} & 58 & 5.0\% & 1 (2\%) & 52 (90\%) & 5 (9\%) \\
    miniF2F & 316 & 18.0\% & 0 (0\%) & 49 (16\%) & 267 (84\%) \\
    \bottomrule
  \end{tabular}
\end{table}

All 16 Babel-formal files are solved at least once, reflecting the controlled tier's tractability. In contrast, only 1 of 58 Formalizing 100 Theorems files has zero passes across all models and directions, but 90\% remain below 10\% pass rate, representing theorems where ecosystem dependencies make translation very difficult for current models. The miniF2F distribution is more spread: most theorems have nonzero pass rates above 10\%, consistent with the statement-only evaluation being more tractable than full proof translation.

\begin{center}
  \centering
  \includegraphics[width=0.55\linewidth]{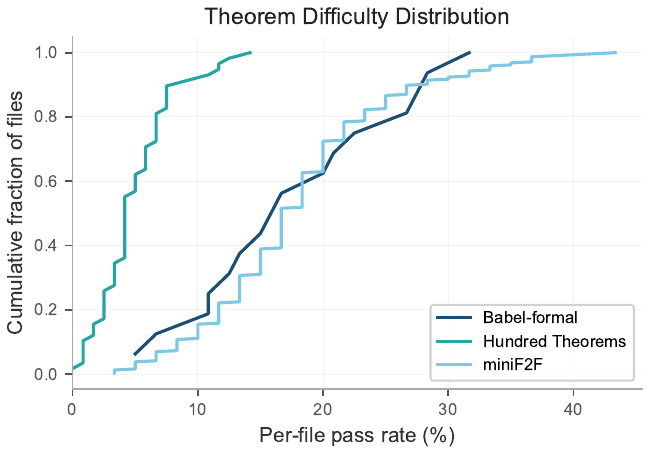}
  \captionof{figure}{Empirical CDF of per-file pass rates by benchmark. Formalizing 100 Theorems files are concentrated near 0\% (90\% below 10\% pass rate), while Babel-formal and miniF2F files spread across a wider range, reflecting their greater tractability.}
  \label{fig:theorem-difficulty-cdf}
\end{center}

\subsection{MiniF2F Competition Categories}
\label{app:minif2f}

MiniF2F theorem titles encode their competition source or mathematical topic. Table~\ref{tab:minif2f-cats} reports statement translation pass rates by category (all models pooled, statement mode only).

\begin{table}[htbp]
  \caption{MiniF2F statement translation pass@1 by competition category (all models pooled).}
  \label{tab:minif2f-cats}
  \centering
  \small
  \begin{tabular}{lccl}
    \toprule
    Category & Passed/Total & \% & Type \\
    \midrule
    MATH Dataset & 2{,}359/12{,}120 & 19.5 & Math dataset \\
    Induction & 127/720 & 17.6 & Topic \\
    AMC 12B & 252/1{,}440 & 17.5 & Competition \\
    AIME & 99/600 & 16.5 & Competition \\
    AMC & 103/660 & 15.6 & Competition \\
    Number Theory & 54/360 & 15.0 & Topic \\
    Algebra & 208/1{,}440 & 14.4 & Topic \\
    AMC 12A & 199/1{,}560 & 12.8 & Competition \\
    IMO & 6/60 & 10.0 & Competition \\
    \bottomrule
  \end{tabular}
\end{table}

\textbf{Competition difficulty does not predict translation difficulty.} Table~\ref{tab:minif2f-hierarchy} groups categories by competition difficulty level. The expected hierarchy AMC $<$ AIME $<$ IMO is \emph{not} reflected in translation pass rates. AIME problems (16.5\%) are slightly easier to translate than AMC problems (14.8\%), and IMO problems (10.0\%) have only a modest gap. These differences should be read as translation statistics for the four-ITP intersection, not as direct measures of mathematical difficulty.

\begin{table}[htbp]
  \caption{MiniF2F pass rates by competition difficulty level. Mathematical difficulty does not predict cross-ITP translation difficulty.}
  \label{tab:minif2f-hierarchy}
  \centering
  \footnotesize
  \setlength{\tabcolsep}{3pt}
  \begin{tabular}{lrrr}
    \toprule
    Difficulty Level & Passed & Total & \% \\
    \midrule
    AMC 12 (incl.\ 12A/B) & 554 & 3660 & 15.1 \\
    AIME & 99 & 600 & 16.5 \\
    IMO / USAMO & 6 & 60 & 10.0 \\
    \bottomrule
  \end{tabular}
\end{table}

Topic-labeled theorems (algebra, number theory, induction: 15.4\% pooled) outperform competition-labeled ones (15.2\%) only marginally in the 4-ITP intersection subset, suggesting that once theorems are formalized across all four systems, topic vs.\ competition provenance has little impact on translation difficulty.

\begin{center}
  \centering
  \includegraphics[width=0.55\linewidth]{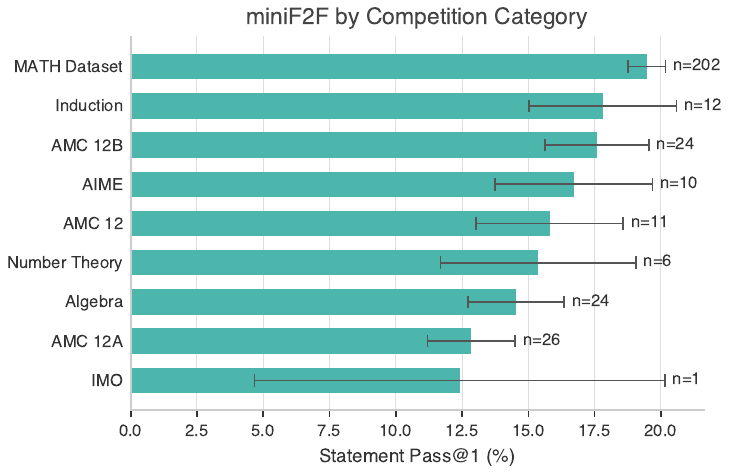}
  \captionof{figure}{MiniF2F statement translation pass@1 by competition category (all models pooled). Mathematical difficulty level (AMC vs.\ AIME vs.\ IMO) does not predict translation difficulty; the MATH Dataset category is easiest to translate despite covering diverse problem types.}
  \label{fig:minif2f-categories}
\end{center}

\FloatBarrier
\section{Extended Related Work}
\label{app:related-work}

\subsection{Formalization tracking and proof exchange}

Several projects study overlap among formal libraries without providing a translation benchmark. Formalizing 100 Theorems~\citep{wiedijk2008hundred} and the 1000+ Theorems project~\citep{thousand2024} catalog which mathematical results have been formalized in which proof assistants. These resources are valuable for identifying common mathematical content, but they do not provide aligned source files, translation tasks, or target prover evaluation.

Logical frameworks and proof exchange systems address a complementary problem: representing proofs from different systems in a common foundation. Dedukti~\citep{dowek2023dedukti} is based on the $\lambda\Pi$-calculus modulo theory and has been used as a target for representing several logics and proof systems. Logipedia~\citep{logipedia2020} and EuroProofNet~\citep{thire2020interoperability} build on this line of work to support proof interoperability. OpenTheory~\citep{opentheory2011} provides a package format and standard library for the HOL family of provers. These projects make proof exchange possible in principle, but they do not evaluate LLM generated translations across many directed prover pairs. \textsc{ITPEval} is therefore complementary: it treats interoperability as an empirical benchmark problem, with target-prover checking as the evaluation signal.

\subsection{Cross ITP translation benchmarks and systems}

Babel-formal~\citep{babel2025} is the closest prior benchmark. It studies Lean and Rocq proof translation using a small aligned corpus of 14 files and 117 lemmas, and explores proof term translation as a bridge between the two systems. \textsc{ITPEval} extends the scope from two systems to four, from one logical family comparison to CIC and HOL, and from 14 aligned files to 390 aligned items (1{,}560 source files across four ITPs).

Recent work on miniF2F in Rocq~\citep{viennot2025minif2f} uses natural language, Lean, and Isabelle sources to produce Rocq theorem statements. This shows that LLMs can help build new formal benchmark ports. However, the task is one target system and statement only. In contrast, \textsc{ITPEval} evaluates all 12 source target directions among Lean~4, Rocq, Isabelle, and HOL Light, and includes both statement and proof translation when aligned proofs exist.

Recent prover agents also include cross system translation as one capability. Leanstral~\citep{leanstral2026}, for example, is designed around Lean~4 and demonstrates Rocq to Lean translation as part of a broader proof engineering agent. Such systems motivate the need for a benchmark that can measure translation quality systematically rather than through isolated examples.

\subsection{Multilingual autoformalization}

Multilingual autoformalization studies translation between informal mathematics and multiple formal languages. MMA~\citep{jiang2023mma} constructs informal formal data from Lean and Isabelle and shows that multilingual training can improve autoformalization. ProofWala~\citep{deshpande2025proofwala} studies multilingual proof data synthesis and cross language transfer between Lean and Rocq. KELPS~\citep{zhang2025kelps} introduces an intermediate representation for translating natural language into Lean, Rocq, and Isabelle.

These works are closely related to our autoformalization and auto-informalization experiment, but differ in direction and evaluation target. They primarily study natural language to formal translation or multilingual training data. \textsc{ITPEval} studies formal-to-formal translation directly and uses natural-language round trips only as a diagnostic for multi-ITP generation context.

\subsection{Semantic fidelity and evaluation}

A recurring issue in autoformalization is that type checking alone can overestimate correctness. \citet{liu2025beq} introduce BEq (Bidirectional Extended Definitional Equivalence), a neuro-symbolic metric that extends definitional equality with tactic-based transformations to determine equivalence between formal statements; we adopt their bidirectional entailment framework for cross-ITP translation evaluation. Poiroux et al.~\citep{poiroux2025reliable} study reliable evaluation for statement autoformalization and introduce stronger equivalence based metrics. Liu et al.~\citep{liu2025gted} propose tree edit based evaluation for formal statements, and ASSESS/TransTED~\citep{assess2026} combines structural and semantic similarity for statement comparison. FormalAlign~\citep{lu2025formalalign} trains models to detect semantic alignment between informal and formal statements, and symbolic equivalence or round trip evaluation methods have also been explored~\citep{li2024symbolic}.

Cross-ITP translation adds another layer of difficulty. The source and target statements often live in different kernels, use different libraries, and expose different definitional equalities. This makes direct structural comparison much harder than in single-prover autoformalization. For this reason, \textsc{ITPEval} uses target-prover checking as the primary executable signal and BEq as a complementary equivalence-aware check for Lean~4 targets.

\subsection{Programmatic prover interfaces}

Machine learning for theorem proving has produced several important prover interaction tools. CoqGym~\citep{yang2019coqgym} provides a dataset and learning environment for Rocq/Coq proof interaction. LeanDojo~\citep{yang2024leandojo} provides Lean data extraction, retrieval augmented proving, and programmatic interaction with Lean. Pantograph~\citep{anand2024pantograph} gives a machine to machine interface for Lean~4 and supports proof search. PISA~\citep{jiang2021pisa} supports interaction with Isabelle. The \texttt{itp-interface} project~\citep{itpinterface2024} provides a generic interface for Lean~4 and Rocq.

These systems focus on proof search or data extraction within one prover, or a small subset of provers. \textsc{ITPEval} instead requires a verification service that can run four heterogeneous target systems at benchmark scale. The main systems challenge is not tactic search, but reproducible target-prover checking under state isolation. Our unified client, filters, environment-keyed scheduling, checkpointed JSONL verification, and warm isolated backends are designed for this cross-prover evaluation setting.

\subsection{Programming language translation benchmarks}

Code translation benchmarks such as CodeTransOcean~\citep{yan2023codetransocean}, HumanEval-X~\citep{zheng2023humaneval}, and CRUXEval-X~\citep{xu2024cruxeval} evaluate translation between programming languages using tests or executable behavior. They are useful analogues, but ITP translation is different in two ways. First, the target artifact must pass a proof assistant kernel, elaborator, tactic engine, and library environment. Second, translation must preserve logical meaning across systems with different foundations and libraries, not merely produce code with the same input output behavior. These differences motivate a dedicated benchmark for formal translation.


\FloatBarrier
\section{Experimental Details}
\label{app:experimental}

\subsection{Autoformalization and Auto-informalization: Full Results}
\label{app:autoformalization}

Table~\ref{tab:autoformalization-full} reports the full multi-ITP round-trip verification results by model and target prover. Rocq and HOL Light account for most verified outputs, but the pattern is not target-independent: GPT-5.5 is strongest on HOL Light and Rocq, Gemini is strongest on Rocq and relatively strong on Isabelle, and the remaining models show smaller or inconsistent gains. Table~\ref{tab:autoformalization-lean-cycle} compares the multi-ITP Lean~4 results against the single-ITP Lean cycle.

\begin{table}[!htbp]
  \caption{Multi-ITP round-trip verification by model and target prover. Entries are pass/checked counts for step~1 and step~3; checked counts exclude empty or skipped outputs.}
  \label{tab:autoformalization-full}
  \centering
  \scriptsize
  \setlength{\tabcolsep}{2.5pt}
  \begin{tabular}{lcccccccc}
    \toprule
    & \multicolumn{2}{c}{Lean~4} & \multicolumn{2}{c}{Isabelle} & \multicolumn{2}{c}{Rocq} & \multicolumn{2}{c}{HOL Light} \\
    \cmidrule(lr){2-3} \cmidrule(lr){4-5} \cmidrule(lr){6-7} \cmidrule(lr){8-9}
    Model & S1 & S3 & S1 & S3 & S1 & S3 & S1 & S3 \\
    \midrule
    GPT-5.5     & 16/100 & 17/100 & 2/100  & 1/100  & 61/100 & 62/100 & 91/100 & 92/100 \\
    Claude 4.6  & 12/100 & 11/100 & 1/100  & 1/100  & 0/100  & 3/100  & 11/100 & 7/100 \\
    Gemini 3.1  & 8/99   & 9/98   & 30/97  & 17/96  & 61/75  & 57/67  & 26/70  & 28/61 \\
    DeepSeek-V4 & 4/94   & 7/96   & 6/94   & 1/95   & 14/92  & 11/94  & 14/91  & 18/93 \\
    Qwen3-235B  & 13/98  & 12/99  & 3/98   & 1/99   & 16/98  & 13/99  & 2/99   & 2/100 \\
    \bottomrule
  \end{tabular}
\end{table}

\begin{table}[!htbp]
  \caption{Lean~4 verification under single-ITP and multi-ITP round trips. Entries count all 100 selected miniF2F theorems per model, treating missing or invalid generations as non-passing.}
  \label{tab:autoformalization-lean-cycle}
  \centering
  \footnotesize
  \setlength{\tabcolsep}{4pt}
  \begin{tabular}{lrrrrrr}
    \toprule
    & \multicolumn{3}{c}{Step 1 pass} & \multicolumn{3}{c}{Step 3 pass} \\
    \cmidrule(lr){2-4} \cmidrule(lr){5-7}
    Model & Single & Multi & $\Delta$ & Single & Multi & $\Delta$ \\
    \midrule
    GPT-5.5     & 2  & 16 & +14 & 4  & 17 & +13 \\
    Claude 4.6  & 13 & 12 & -1  & 15 & 11 & -4 \\
    Gemini 3.1  & 0  & 8  & +8  & 0  & 9  & +9 \\
    DeepSeek-V4 & 4  & 4  & 0   & 5  & 7  & +2 \\
    Qwen3-235B  & 5  & 13 & +8  & 15 & 12 & -3 \\
    \midrule
    Pooled      & 24 & 53 & +29 & 39 & 56 & +17 \\
    \bottomrule
  \end{tabular}
\end{table}

\paragraph{Worked example.}
We trace the full multi-ITP round trip for AIME 1983 Problem~1 (Claude Sonnet 4.6) to illustrate the process and a characteristic failure mode. The original NL statement is:

\begin{promptbox}[Step 0: Reference NL]
\small\ttfamily
Let x, y and z all exceed 1 and let w be a positive number such that log\_x w = 24, log\_y w = 40 and log\_\{xyz\} w = 12. Find log\_z w. Show that it is 060.
\end{promptbox}

\noindent\textbf{Step~1} (NL $\mapsto$ 4 ITPs) produces four formalizations. Two representative outputs:

\begin{promptbox}[Step 1: NL $\mapsto$ Lean 4]
\small\ttfamily
theorem log\_z\_w (x y z w : $\mathbb{R}$) (hx : x > 1) (hy : y > 1) (hz : z > 1) (hw : w > 0)\\
\quad (h1 : Real.log w / Real.log x = 24)\\
\quad (h2 : Real.log w / Real.log y = 40)\\
\quad (h3 : Real.log w / Real.log (x * y * z) = 12) :\\
\quad Real.log w / Real.log z = 60 := by sorry
\end{promptbox}

\begin{promptbox}[Step 1: NL $\mapsto$ Isabelle]
\small\ttfamily
theorem log\_z\_w:\\
\quad fixes x y z w :: real\\
\quad assumes hx: "x > 1" and hy: "y > 1" and hz: "z > 1" and hw: "w > 0"\\
\quad assumes h1: "log x w = 24"\\
\quad assumes h2: "log y w = 40"\\
\quad assumes h3: "log (x * y * z) w = 12"\\
\quad shows "log z w = 60"\\
\quad sorry
\end{promptbox}

\noindent\textbf{Step~2} (4 ITPs $\mapsto$ NL) produces an informalized description from all four formalizations:

\begin{promptbox}[Step 2: 4 ITPs $\mapsto$ NL]
\small\ttfamily
If x, y, z > 1 and w > 0, and log\_x(w) = 24, log\_y(w) = 40, and log\_\{xyz\}(w) = 12, then log\_z(w) = 60.
\end{promptbox}

\noindent\textbf{Step~3} (NL $\mapsto$ 4 ITPs again) re-formalizes from the step-2 NL. The Lean~4 output:

\begin{promptbox}[Step 3: NL $\mapsto$ Lean 4 (second round)]
\small\ttfamily
theorem log\_problem (x y z w : $\mathbb{R}$) (hx : x > 1) (hy : y > 1) (hz : z > 1) (hw : w > 0)\\
\quad (h1 : Real.log w / Real.log x = 24)\\
\quad (h2 : Real.log w / Real.log y = 40)\\
\quad (h3 : Real.log w / Real.log (x * y * z) = 12) :\\
\quad Real.log w / Real.log z = 60 := by sorry
\end{promptbox}

The step-1 and step-3 Lean~4 outputs are syntactically close and express the same real-valued statement (modulo the theorem name), illustrating round-trip stability for this theorem. However, a subtle semantic shift occurs relative to the \emph{reference} formalization: the original miniF2F Lean~4 statement types $x, y, z$ as $\mathbb{N}$ (natural numbers) with $w : \mathbb{N}$, while the round-trip-generated versions use $\mathbb{R}$ (reals), yielding a strictly weaker statement. This information loss is introduced at step~1, where the NL description does not specify integrality, and propagates unchanged through the rest of the round trip. Such type-level semantic drift, invisible in the NL domain, is a characteristic failure mode of NL-mediated round-trips.

\subsection{Prompt Templates}
\label{app:prompts}

All evaluations use a two-message format (system + user) with zero-shot prompting. Placeholders in \texttt{\{braces\}} are filled per translation pair. The target-language header at the end of each prompt primes generation with the correct output format. System messages instruct the model to return only raw ITP code with no markdown fencing or explanation.

\begin{promptbox}[Statement translation]
\small\ttfamily
Translate the following \{src\_name\} theorem statement of "\{title\}" into \{tgt\_name\}. Do not prove the theorem; leave the proof body as sorry or the equivalent placeholder.

\medskip
=== \{src\_name\} statement ===\\
\{source\_content\}

\medskip
=== \{tgt\_name\} statement ===
\end{promptbox}

\begin{promptbox}[Proof translation]
\small\ttfamily
Translate the following \{src\_name\} proof of "\{title\}" into \{tgt\_name\}.

\medskip
=== \{src\_name\} proof ===\\
\{source\_content\}

\medskip
=== \{tgt\_name\} translation ===
\end{promptbox}

\begin{promptbox}[Round-trip: NL $\mapsto$ Lean~4]
\small\ttfamily
Translate the following natural language theorem statement into Lean 4. Do not prove the theorem; leave the proof body as sorry.

\medskip
=== Natural language statement ===\\
\{nl\}

\medskip
=== Lean 4 statement ===
\end{promptbox}

\begin{promptbox}[Round-trip: Lean~4 $\mapsto$ NL]
\small\ttfamily
Describe the following Lean 4 theorem statement in natural language. Give a concise mathematical statement of what the theorem claims.

\medskip
=== Lean 4 statement ===\\
\{lean4\}

\medskip
=== Natural language description ===
\end{promptbox}

\begin{promptbox}[Round-trip: NL $\mapsto$ 4~ITPs]
\small\ttfamily
Translate the following natural language theorem statement into formal statements in all four ITPs. Leave proof bodies as sorry or the equivalent placeholder in each system.

\medskip
=== Natural language statement ===\\
\{nl\}

\medskip
=== Lean 4 ===

=== Isabelle ===

=== Rocq ===

=== HOL Light ===
\end{promptbox}

\begin{promptbox}[Round-trip: 4~ITPs $\mapsto$ NL]
\small\ttfamily
The following formal theorem statements all describe the same mathematical theorem. Give a concise natural language description of what the theorem claims.

\medskip
=== Lean 4 ===\\
\{lean4\}

=== Isabelle ===\\
\{isabelle\}

=== Rocq ===\\
\{rocq\}

=== HOL Light ===\\
\{hol\_light\}

\medskip
=== Natural language description ===
\end{promptbox}

\subsection{Babel-formal Formalization Process}
\label{app:babel-formalization}

The 16 Babel-formal files originate in Lean~4 and Rocq (adopted from the Babel-formal benchmark~\citep{babel2025}, plus 2 new files). To complete the four-way alignment required by our benchmark design, we manually formalized all 16 files in Isabelle and HOL Light. This section describes the translation methodology, key design decisions, and illustrative examples.

\paragraph{Design principle: self-contained axiomatization.}
Every Babel-formal file is designed to be self-contained: it carries its own definitions and axioms rather than depending on any ITP-specific library. This property is inherited from the original Lean/Rocq sources and is preserved in the Isabelle and HOL Light translations. This axiomatized design serves two purposes: (1)~it ensures that translation difficulty reflects only foundational differences (type systems, proof languages, abstraction mechanisms) rather than library API mismatches, and (2)~it makes the files independently verifiable without installing large library ecosystems.

\paragraph{Illustrative example: \texttt{comp\_commute}.}
We show the complete statement-mode file for \texttt{comp\_commute} (function composition commutativity) in all four ITPs.\footnote{The actual source files use Unicode Greek letters (\texttt{$\alpha$}, \texttt{$\beta$}, \texttt{$\gamma$}) for type variables in Lean~4 and Rocq, and Unicode symbols (\texttt{$\Rightarrow$}, \texttt{$\Longrightarrow$}, \texttt{$\equiv$}, \texttt{$\lambda$}) in Isabelle. We render these as ASCII equivalents for typographic consistency.} This file defines composition and identity, then states seven lemmas about commutativity with proof bodies replaced by placeholders, exactly the input an LLM receives during statement translation. The four versions illustrate how the same mathematical content is expressed differently across foundations:

\begin{codebox}[Lean 4]
universe u v w

namespace CompCommute

variable {a : Type u} {b : Type v} {c : Type w}

def comp {a b c} (g : b -> c) (f : a -> b) : a -> c := fun x => g (f x)
def id {a} : a -> a := fun x => x

axiom comp_assoc :
  forall {a b c d} (h : c -> d) (g : b -> c) (f : a -> b),
  comp h (comp g f) = comp (comp h g) f
axiom comp_id_l : forall {a b} (f : a -> b), comp (id) f = f
axiom comp_id_r : forall {a b} (f : a -> b), comp f id = f

def commute {a} (f g : a -> a) : Prop := comp f g = comp g f

theorem commute_symm {a} (f g : a -> a) :
  commute f g -> commute g f := by sorry
theorem commute_with_id_l {a} (f : a -> a) : commute f (id) := by sorry
theorem commute_with_id_r {a} (f : a -> a) : commute (id) f := by sorry
theorem commute_refl {a} (f : a -> a) : commute f f := by sorry
theorem commute_congr {a} (f1 f2 g1 g2 : a -> a) :
  f1 = f2 -> g1 = g2 -> commute f1 g1 -> commute f2 g2 := by sorry
theorem commute_transport_left_id {a} (f g : a -> a) :
  commute f g -> commute (comp (id) f) g := by sorry
theorem commute_transport_right_id {a} (f g : a -> a) :
  commute f g -> commute f (comp (id) g) := by sorry
end CompCommute
\end{codebox}

\begin{codebox}[Rocq]
Section CompCommute.
  Universes u v w.
  Definition comp {a b c : Type} (g : b -> c) (f : a -> b)
    : a -> c := fun x => g (f x).
  Definition id {a : Type} : a -> a := fun x => x.

  Class FunProps := {
    comp_assoc : forall {a b c d} (h : c -> d) (g : b -> c)
      (f : a -> b),
      comp h (comp g f) = comp (comp h g) f;
    comp_id_l : forall {a b} (f : a -> b),
      comp (@id b) f = f;
    comp_id_r : forall {a b} (f : a -> b),
      comp f (@id a) = f
  }.

  Definition commute {a : Type} (f g : a -> a) : Prop :=
    comp f g = comp g f.
  Context {FP : FunProps}.

  Lemma commute_symm :
    forall {a} (f g : a -> a), commute f g -> commute g f.
  Proof. Admitted.
  Lemma commute_with_id_l :
    forall {a} (f : a -> a), commute f (@id a).
  Proof. Admitted.
  Lemma commute_with_id_r :
    forall {a} (f : a -> a), commute (@id a) f.
  Proof. Admitted.
  Lemma commute_refl :
    forall {a} (f : a -> a), commute f f.
  Proof. Admitted.
  Lemma commute_congr :
    forall {a} (f1 f2 g1 g2 : a -> a),
    f1 = f2 -> g1 = g2 ->
    commute f1 g1 -> commute f2 g2.
  Proof. Admitted.
  Lemma commute_transport_left_id :
    forall {a} (f g : a -> a),
    commute f g -> commute (comp (@id a) f) g.
  Proof. Admitted.
  Lemma commute_transport_right_id :
    forall {a} (f g : a -> a),
    commute f g -> commute f (comp (@id a) g).
  Proof. Admitted.
End CompCommute.
\end{codebox}

\begin{codebox}[Isabelle]
theory comp_commute
  imports Main
begin

definition myComp ::
    "('b => 'c) => ('a => 'b) => ('a => 'c)"
  where "myComp g f == (%x. g (f x))"
definition myId :: "'a => 'a"
  where "myId == (%x. x)"

lemma comp_assoc:
  "myComp h (myComp g f) = myComp (myComp h g) f"
  by sorry
lemma comp_id_l: "myComp myId f = f" by sorry
lemma comp_id_r: "myComp f myId = f" by sorry

definition commute ::
    "('a => 'a) => ('a => 'a) => bool"
  where "commute f g == myComp f g = myComp g f"

lemma commute_symm:
  "commute f g ==> commute g f" by sorry
lemma commute_with_id_l: "commute f myId" by sorry
lemma commute_with_id_r: "commute myId f" by sorry
lemma commute_refl: "commute f f" by sorry
lemma commute_congr:
  "f1 = f2 ==> g1 = g2 ==>
   commute f1 g1 ==> commute f2 g2"
  by sorry
lemma commute_transport_left_id:
  "commute f g ==>
   commute (myComp myId f) g"
  by sorry
lemma commute_transport_right_id:
  "commute f g ==>
   commute f (myComp myId g)"
  by sorry
end
\end{codebox}

\begin{codebox}[HOL Light]
let myComp = new_definition
  `myComp (g:B->C) (f:A->B) = (\x:A. g (f x))`;;
let myId = new_definition
  `myId = (\x:A. x)`;;

let comp_assoc =
  `!h g f. myComp h (myComp g f)
           = myComp (myComp h g) f`;;
let comp_id_l = `!f:A->B. myComp myId f = f`;;
let comp_id_r = `!f:A->B. myComp f myId = f`;;

let commute = new_definition
  `commute (f:A->A) (g:A->A) <=>
   myComp f g = myComp g f`;;

let commute_symm =
  `!f g. commute f g ==> commute g f`;;
let commute_with_id_l =
  `!f:A->A. commute f myId`;;
let commute_with_id_r =
  `!f:A->A. commute myId f`;;
let commute_refl = `!f:A->A. commute f f`;;
let commute_congr =
  `!f1 f2 g1 g2:A->A.
    f1 = f2 ==> g1 = g2 ==>
    commute f1 g1 ==> commute f2 g2`;;
let commute_transport_left_id =
  `!f g:A->A.
    commute f g ==>
    commute (myComp myId f) g`;;
let commute_transport_right_id =
  `!f g:A->A.
    commute f g ==>
    commute f (myComp myId g)`;;
\end{codebox}

Several differences are visible even in this simple file. Lean~4 uses universe-polymorphic type variables (\texttt{\{$\alpha$ : Type u\}}) with implicit arguments; Rocq bundles the function axioms into a typeclass (\texttt{Class FunProps}) and threads the instance via \texttt{Context}; Isabelle redefines composition and identity as \texttt{myComp}/\texttt{myId} (avoiding name clashes with the standard library's \texttt{o} and \texttt{id}); and HOL Light states lemmas as bare quoted terms (not wrapped in \texttt{prove}) since statement-mode files contain no proofs. The placeholder mechanisms also differ: \texttt{sorry} (Lean/Isabelle), \texttt{Admitted} (Rocq), and omission of \texttt{prove} (HOL Light).

For files with algebraic structure hierarchies (e.g., \texttt{group}, \texttt{ideals}, \texttt{lattice\_like}), the differences are more pronounced: Lean/Rocq typeclasses become Isabelle locales (e.g., \texttt{locale group = fixes mul ...~assumes mul\_assoc ...}) and HOL Light parameterized predicates (e.g., \texttt{is\_group (mul:A->A->A) (e:A) (ginv:A->A) <=> ...}). Lean/Rocq class inheritance (e.g., \texttt{GroupComm} extending \texttt{Group}) becomes locale composition (\texttt{locale group\_comm = group +}) in Isabelle and predicate conjunction in HOL Light.

\paragraph{Proof translation strategies.}
Proof styles differ across ITPs.
\emph{Isabelle}: We wrote all proofs as structured Isar proofs, using \texttt{calc} chains for equational reasoning and Isabelle's automation (\texttt{simp}, \texttt{metis}, \texttt{blast}) for routine steps.
\emph{HOL Light}: For simpler files (e.g., \texttt{comp\_commute}, \texttt{group}), we wrote complete tactic proofs. For the \texttt{group} file, a key technique is the ``babel group'' bridge: we embed the parameterized group into HOL Light's native \texttt{group} type via a definitional bridge, allowing reuse of the standard \texttt{grouptheory.ml} library lemmas while maintaining the axiomatized interface.

For the most complex files (notably \texttt{polynomial}), we adopted an \emph{axiomatic} strategy in HOL Light: benchmark lemmas are stated as axioms (\texttt{new\_axiom}) rather than derived. This is a benchmark engineering choice rather than a proof-exchange claim: (1)~the corresponding mathematical development is checked in Lean~4, Rocq, and Isabelle; (2)~HOL Light's lack of locale/typeclass abstractions makes certain proof structures (e.g., nested ring--polynomial locale contexts with 30+ parameters) prohibitively verbose to replicate; and (3)~the axioms still serve the benchmark's purpose since the HOL Light file is used only as a \emph{source} or \emph{target} for LLM translation, not as a trust root. Definitions and helper predicates (e.g., \texttt{is\_poly\_context}, which bundles 30+ ring and polynomial axioms into a single predicate) are still fully specified, ensuring that name resolution, type checking, and file structure are exercised by the verifier.

\paragraph{Summary statistics.}
Across the 16 files (2{,}902 lines in Lean~4; 2{,}725 in Rocq), the Isabelle translations total 2{,}966 lines, nearly isomorphic to the originals, with the slight increase due to locale declaration syntax and explicit \texttt{imports} headers. The HOL Light translations total 2{,}407 lines. Although they are shorter overall, individual lemma signatures are longer because they use explicit universal quantification over structure parameters and the \texttt{is\_poly\_context} bundled-predicate pattern; the axiomatic strategy used for the most complex files reduces proof-side line count. All 16 Isabelle and HOL Light files compile cleanly in Isabelle~2024 and HOL Light (latest git), respectively.

\section{Semantic Equivalence Checking Details}
\label{app:beq}

This appendix describes the BEq protocol, provides per-model directional breakdowns, documents entailment asymmetry, and gives examples of verified-but-semantically-incorrect translations.

\subsection{BEq Protocol}
\label{app:beq-protocol}

The BEq check tests whether a verified translation preserves the mathematical meaning of the reference theorem. In this paper, BEq is applied to verified Lean~4 target translations. Given a generated Lean statement $G$ and the reference Lean statement $R$, BEq constructs two entailment problems:

\begin{itemize}[leftmargin=*]
    \item \textbf{Forward} ($G \vdash R$): Assume $G$ as an axiom; attempt to prove $R$.
    \item \textbf{Backward} ($R \vdash G$): Assume $R$ as an axiom; attempt to prove $G$.
\end{itemize}

A translation passes BEq iff both directions succeed. Forward-only means the generated statement is \emph{stronger} than the reference; backward-only means it is \emph{weaker}.

\paragraph{Lean construction.}
Each directional check is compiled as an independent Lean file:

\begin{enumerate}[leftmargin=*, label=\arabic*.]
    \item Split the reference file into a header and theorem block; the reference header supplies the Lean imports, \texttt{open} commands, and options used for the check.
    \item Extract the generated theorem block from the verified model output.
    \item Rename the assumed theorem to \texttt{stmt\_assumed}.
    \item Rename the goal theorem to \texttt{beq\_goal}.
    \item Replace the goal proof with a restricted proof search and run Lean.
\end{enumerate}

\paragraph{Deterministic restricted search.}
\citet{liu2025beq} define BEq through a restricted transformation function: first try Lean's \texttt{exact?}, then sample tactic sequences from an LLM under a restricted primitive set. Their ablations show that bidirectionality is important and that restricting the primitive set controls false positives. We preserve this design principle but make the transformation search deterministic. The check first runs \texttt{exact?}; the attempt is accepted only if Lean's suggestion explicitly references \texttt{stmt\_assumed}. If this fails, the checker tries tactics corresponding to the normal primitive set: \texttt{exact}, \texttt{apply}, \texttt{rw}, reverse rewriting, \texttt{intro}/\texttt{intros}, \texttt{constructor}, \texttt{ext}, local \texttt{have}, case splitting, and existential \texttt{use}. Each tactic is instantiated so that a successful proof must use \texttt{stmt\_assumed}, preventing standalone automation from proving the goal independently of the assumed statement.

\paragraph{Interpretation.}
BEq is a Lean-checked but incomplete equivalence test. A pass is witnessed by Lean proofs of both entailments. A failure means only that the deterministic restricted search did not establish equivalence; logically equivalent statements can still fail when they require transformations outside the implemented cascade, exceed Lean's search limits, or are not extracted into the expected theorem shape.

\subsection{Per-Model Directional Breakdown}
\label{app:beq-direction}

Table~\ref{tab:beq-direction} reports BEq pass rates broken down by model and source prover.

\begin{table}[!htbp]
  \caption{BEq pass rate by model and source $\mapsto$ Lean~4 direction (pass/verified).}
  \label{tab:beq-direction}
  \centering
  \tiny
  \setlength{\tabcolsep}{2pt}
  \begin{tabular}{lcccr}
    \toprule
    Model & Rocq $\mapsto$ Lean & Isa $\mapsto$ Lean & HOL $\mapsto$ Lean & Total \\
    \midrule
    Claude Sonnet 4.6  & 36/43 (83.7\%) & 1/3 (33.3\%) & 25/28 (89.3\%) & 62/74 (83.8\%) \\
    DeepSeek-V4-Pro    & 16/42 (38.1\%) & 7/18 (38.9\%) & 14/24 (58.3\%) & 37/84 (44.0\%) \\
    Gemini 3.1 Pro     & 3/7 (42.9\%)   & --- & --- & 3/7 (42.9\%) \\
    GPT-5.5            & 5/20 (25.0\%) & --- & 5/9 (55.6\%) & 10/29 (34.5\%) \\
    Qwen3-235B         & 29/58 (50.0\%) & 21/40 (52.5\%) & 12/30 (40.0\%) & 62/128 (48.4\%) \\
    \midrule
    All models         & 89/170 (52.4\%) & 29/61 (47.5\%) & 56/91 (61.5\%) & 174/322 (54.0\%) \\
    \bottomrule
  \end{tabular}
\end{table}

BEq pass rates vary by model and source direction. Claude Sonnet~4.6 is strongest overall, while GPT-5.5 and DeepSeek-V4-Pro have low BEq rates despite producing verified Lean statements. Across directions, HOL Light $\mapsto$ Lean has the highest aggregate BEq rate (61.5\%), followed by Rocq $\mapsto$ Lean (52.4\%) and Isabelle $\mapsto$ Lean (47.5\%).

\subsection{Forward vs.\ Backward Entailment}
\label{app:beq-symmetry}

Table~\ref{tab:beq-symmetry} reports the breakdown of entailment directions among verified translations.

\begin{table}[!htbp]
  \caption{Entailment direction breakdown for verified source $\mapsto$ Lean~4 translations. ``Both'' = BEq pass; ``BwdOnly'' = reference entails generated but not vice versa (weaker translation).}
  \label{tab:beq-symmetry}
  \centering
  \scriptsize
  \setlength{\tabcolsep}{3pt}
  \begin{tabular}{lrrrrr}
    \toprule
    Model & Checked & Both & Neither & FwdOnly & BwdOnly \\
    \midrule
    Claude Sonnet 4.6 & 74 & 62 & 10 & 0 & 2 \\
    DeepSeek-V4-Pro & 84 & 37 & 40 & 0 & 7 \\
    Gemini 3.1 Pro & 7 & 3 & 4 & 0 & 0 \\
    GPT-5.5 & 29 & 10 & 18 & 0 & 1 \\
    Qwen3-235B & 128 & 62 & 60 & 0 & 6 \\
    \midrule
    All models & 322 & 174 & 132 & 0 & 16 \\
    \bottomrule
  \end{tabular}
\end{table}

All 16 asymmetric cases are backward-only; no model produces any forward-only cases (translations stronger than the reference). This pattern is consistent with \emph{weakening}: dropping hypotheses, loosening types, or shifting to a more general formulation that the reference can still imply. Most failures, however, fail in both directions (132 cases). Additionally, 26 verified records have extraction errors and are counted as BEq failures in the aggregate results.

\subsection{Failure Examples}
\label{app:beq-examples}

We illustrate three classes of BEq failures from DeepSeek-V4-Pro's backward-only cases.

\paragraph{Genuine semantic weakening.}
Theorem \texttt{amc12a\_2021\_p3} asks to show $x - y = 14238$ given constraints on $x, y$. The reference uses integer coercion (\texttt{$\uparrow$x~-~$\uparrow$y~=~(14238:\ensuremath{\mathbb{Z}})}), while DeepSeek generates natural-number subtraction (\texttt{x~-~y~=~14238}). Since $\mathbb{N}$ subtraction truncates at zero, the generated statement is strictly weaker. The reference entails the generated (backward), but not vice versa (forward).

\paragraph{Definitional equivalence beyond the cascade.}
Theorem \texttt{amc12a\_2003\_p5} encodes digit constraints. The reference uses \texttt{Nat.ofDigits 10 [0,1,C,M,A]}, while DeepSeek inlines the arithmetic as \texttt{10000*A + 1000*M + 100*C + 10*1 + 0}. These are definitionally equal in Lean~4, but the tactic cascade cannot close the forward direction. This is a false negative: the statements are logically equivalent, but BEq reports backward-only.

\paragraph{Structural reformulation.}
Theorem \texttt{amc12\_2000\_p1} has the reference state hypotheses as separate binder arguments (\texttt{(h$_0$ : i $\neq$ m $\wedge$ ...) (h$_1$ : i*m*o = 2001)}), while DeepSeek bundles everything into a single implication (\texttt{(i $\neq$ m $\wedge$ ... $\wedge$ i*m*o = 2001) $\to$ ...}). The two formulations are logically equivalent, but the cascade hits a recursion depth limit on the forward direction. This is another false negative from tactic limitations.

\paragraph{Limitations.}
The tactic cascade is deliberately conservative: it uses only standard Lean~4 automation and does not attempt custom proof search. This means that some backward-only and neither-direction cases may be logically equivalent to the reference but beyond the cascade's reach. Prototype extensions to Rocq (using \texttt{auto}, \texttt{eauto}, \texttt{lia}, etc.), Isabelle (using \texttt{auto}, \texttt{blast}, \texttt{sledgehammer}, etc.), and HOL Light (using \texttt{MESON\_TAC}, \texttt{ARITH\_TAC}, etc.) exist, but this paper reports only the Lean~4 deployment; evaluating non-Lean BEq at scale remains future work.

\FloatBarrier
\section{Infrastructure Details}
\label{app:infra}

Section~\ref{sec:infra} rests on one invariant: every accelerated backend must be observationally equivalent to checking the generated artifact in isolation. Concatenating generated files into one prover session is unsound for this benchmark because later artifacts could depend on earlier declarations and top-level-name collisions could turn otherwise valid artifacts into failures.

\paragraph{Toolchain manager.}
The released \texttt{itpeval} CLI installs and runs proof assistants under a common prefix. Each prover adapter exposes installation and checking entry points, and \texttt{run\_itp()} materializes a prover-specific task directory before invoking the native binary. The adapter returns a structured result containing exit status, wall-clock time, captured logs, and backend metadata. Adding a new prover requires only an adapter that maps generated source text to a native check task and returns this result object.

\paragraph{State-isolated resident verification.}
Cold checking is simple but too slow for large cross-ITP evaluation. The resident layer therefore amortizes startup while preserving per-file semantics. Lean requests are fresh synthetic files, Rocq requests are followed by \texttt{Reset Initial}, Isabelle requests are temporary theories in server sessions, and HOL Light requests use either collision-aware worker packing or a forkserver whose child exits after each request.

\begin{table}[htbp]
  \caption{Resident backend semantics. Each backend amortizes setup cost while preserving one generated artifact as the evaluation unit.}
  \label{tab:backend-semantics}
  \centering
  \scriptsize
  \setlength{\tabcolsep}{2pt}
  \begin{tabular}{llll}
    \toprule
    Prover & Backend & Reuse & Isolation \\
    \midrule
    Lean~4 & pool/server & compiler workers & fresh file \\
    Rocq & REPL pool & persistent proc. & \texttt{Reset Initial.} \\
    Isabelle & server-batch & persistent session & temp.\ theories \\
    HL & resident pool & preloaded OCaml & collision packing \\
    HL & forkserver & preloaded parent & child exit \\
    \bottomrule
  \end{tabular}
\end{table}

\paragraph{Scheduling, caching, and bisection.}
Each generated record is assigned an environment key consisting of the target prover, benchmark mode, imports, and backend constraints. The scheduler groups compatible records, orders groups by estimated cost, and dispatches them to the fastest backend that preserves per-file semantics. Strict filters are deliberately one-sided: they can reject invalid artifacts, such as proof-mode outputs containing placeholders, but they can never turn an artifact into a success. For batch-capable backends, failures are localized by adaptive bisection until per-record labels are recovered.

\paragraph{Checkpointing and telemetry.}
Verification writes append-only JSONL checkpoints keyed by stable record identifiers, so interrupted runs resume without losing completed labels. Each checked record stores the original generated text, pass/fail label, first diagnostic, wall-clock time, backend name, environment key, scheduling metadata, and cache status. This telemetry distinguishes model failures from verifier or toolchain failures and enables cold-equivalence audits of resident backends.

\begin{center}
  \centering
  \includegraphics[width=\linewidth]{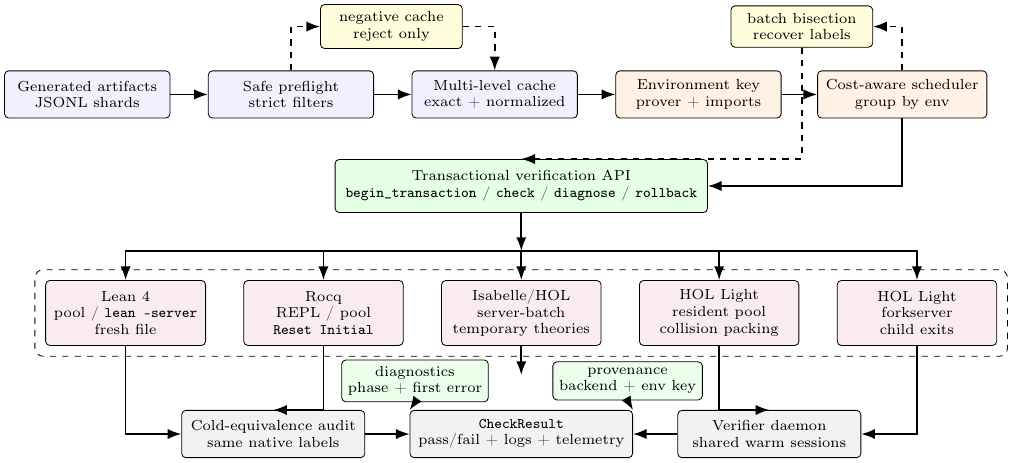}
  \captionof{figure}{Expanded infrastructure view for the state-isolated verification service. The same architecture is summarized in the main text.}
  \label{fig:infra-architecture}
\end{center}

\subsection{Reproducibility}
\label{app:reproduce}

All code, data, and evaluation artifacts are available in the public \texttt{itpeval} release. The repository includes: (1)~the complete benchmark data in structured JSON format (1{,}560 statement files and 296 proof files across four ITPs), (2)~the evaluation pipeline with both batch and synchronous execution modes supporting all five models, (3)~bootstrap scripts that install all four ITP toolchains on macOS and Linux, (4)~all raw model outputs and native ITP verification results, and (5)~scripts to reproduce all tables and figures in this paper.

The evaluation pipeline is designed for end-to-end reproducibility. LLM translation and ITP verification are decoupled: translations can be generated on any machine with API keys, while verification requires the ITP toolchains and can run separately on the same pre-generated outputs. All runs produce checkpointed JSONL files that support safe interruption and resumption. The pipeline is also extensible: new models can be added with a single configuration line, new providers by implementing a dispatch function, and new benchmark data by appending records to the data files (every theorem must have all four ITP variants).

\paragraph{Compute resources.}
All LLM generation was performed through provider APIs (OpenAI, Anthropic, Google, OpenRouter); no local GPU compute was required. ITP verification is CPU-only and runs on any machine with the four toolchains installed. Per-record verification telemetry (wall-clock time, backend provenance, cache status) is included in the released JSONL checkpoints.

The autoformalization round-trip experiment uses a curated 100-record subset of miniF2F, selected to avoid the test split's ${\sim}62\%$ \texttt{mathd} dominance (all 60 non-\texttt{mathd} test records plus 40 randomly sampled \texttt{mathd} records, seed=42). Both single-ITP (NL $\mapsto$ Lean~4 $\mapsto$ NL $\mapsto$ Lean~4) and multi-ITP (NL $\mapsto$ 4 ITPs $\mapsto$ NL $\mapsto$ 4 ITPs) cycles are included.

\newpage

\end{document}